%% file: T-SP-topoid_directed.tex
\documentclass[10pt,twocolumn,twoside]{IEEEtran}

\usepackage{tcolorbox}
\usepackage{graphicx}
\usepackage{xcolor}
\usepackage{theorem}
\usepackage{times,amsmath,epsfig}
\usepackage{amssymb}
\usepackage{cite}
\usepackage{cases}
\usepackage{url}
\usepackage{multicol}
\usepackage{subcaption}
\usepackage{booktabs}

\usepackage{algorithmic}
\usepackage{algorithm}
\usepackage[normalem]{ulem}
\input{my_sections.tex}

\usepackage{tikz}
\usetikzlibrary{shapes,arrows}
\input{mysymbol.sty}

\newcommand{\EE}{{\mathbb E}}

\DeclareMathOperator{\grad}{grad}

\newtheorem{mytheorem}{\bf Theorem}

\newtheorem{mycorollary}{\bf Corollary}
\newtheorem{mylemma}{\bf Lemma}
\newtheorem{myproposition}{\bf Proposition}
\newtheorem{remark}{\bf Remark}

\newtcolorbox{myblockt}[1]{colback=urblue!5!white,
	colframe=urblue,fonttitle=\bfseries,
	title=#1}

\newtcolorbox{myblock}{colback=urblue!5!white,
	colframe=urblue,fonttitle=\bfseries}

\title{Directed Graph Topology Inference via\\ Graph Filter Identification}

\author{\IEEEauthorblockN{Rasoul Shafipour$^\ast$, Andrei Buciulea$^\ast$, \emph{Member, IEEE}, Santiago Segarra, \emph{Senior Member, IEEE,}\\  Antonio G. Marques, \emph{Senior Member, IEEE,} and Gonzalo Mateos, \emph{Senior Member, IEEE}}
\thanks{$^\ast$The first two authors contributed equally. Work in this paper was supported in part by the NSF awards CCF-1750428 and CCF-2340481; the Spanish AEI (AEI/10.13039/501100011033) grant PID2022-136887NB-I00; and the Community of Madrid via IDEA-CM (TEC-2024/COM-89) and the Ellis Madrid Unit. R. Shafipour is with NVIDIA. A. Buciulea and A. G. Marques are with the Dept. of Signal Theory and Comms., King Juan Carlos University. S. Segarra is with the Dept. of Electrical and Computer Eng., Rice University. G. Mateos is with the Dept. of Electrical and Computer Eng., University of Rochester. Emails:
rashafipour@gmail.com,
andrei.buciulea@urjc.es, segarra@rice.edu, antonio.garcia.marques@urjc.es, and gmateosb@ece.rochester.edu. Part of the results in this paper were presented at the \textit{2018 IEEE Data Science Workshop}~\cite{RSSSAMGM_dsw18}.}}

\begin{document}
\maketitle

\begin{abstract}%
We address the problem of inferring a \textit{directed} network from nodal measurements  generated by linear diffusion dynamics on the sought graph. Observations are modeled as the outputs of a graph convolutional filter, i.e., a polynomial (with unknown coefficients) of a local diffusion graph-shift operator encoding the latent graph topology, excited with an ensemble of independent graph signals with arbitrarily-correlated nodal components. Unlike prior efforts that considered undirected graphs and white signal excitations, here the graph-shift operator and the observations' covariance matrix are not simultaneously diagonalizable. In this challenging context, we first rely on measurements of the output signals along with prior statistical information on the inputs to identify the diffusion filter. Such system identification problem involves solving a system of quadratic matrix equations, which we show is identifiable under spectral-diversity assumptions on the input covariances. For algorithmic purposes we recast it as a smooth quadratic minimization subject to Stiefel manifold constraints. Subsequent identification of the network topology given the graph filter estimate boils down to finding a sparse and structurally admissible shift that commutes with the given filter, thus, forcing the latter to be a polynomial in the sought graph-shift operator. A joint graph filter and topology identification algorithm is also proposed, which alternates between the aforementioned steps in a mutually reinforcing fashion to offer improved sample complexity. Numerical tests corroborate the effectiveness of the proposed algorithms in recovering synthetic digraphs and real-data case studies, and illustrate their potential utility on urban mobility analyses as well as portfolio optimization.
\end{abstract}


%
\section{Introduction}\label{S:Introduction}

Machine learning and signal processing over graphs offer a rich toolbox for extracting actionable information from relational data. Upon modeling the data domain as a graph and the nodal observations at hand as graph signals, the graph signal processing (GSP) body of work has contributed innovative models that relate the properties of the signals with those of the graph, along with algorithms that fruitfully leverage graph-induced priors and inductive biases. Arguably, most early GSP efforts dealt with \textit{undirected} networks. Such graphs are equivalently represented by symmetric matrices whose (well-behaved) spectral properties can be used to define transforms, convolutional operators, and even non-linear parametric architectures to learn from graph signals~\cite{ortega2018pieee,gama2020spmag}. Their scarcer adoption notwithstanding, \emph{directed graph} (digraph) models are more adequate (and, in fact, more accurate) for a number of applications. Information networks such as scientific citations or the Web itself are typically directed, and flows in technological (e.g., transportation, power, communication) networks are oftentimes one-directional. The presence of directionality plays a critical role when the measurements taken in those networks need to be processed to remove noise, outliers and artifacts, and this requires new tools and algorithms that do not rely on symmetric graph operators~\cite{SandryMouraSPG_TSP13,DG_SPMAG}. More abstractly, when the graph encodes (often unknown) relations between observed variables, directionality is vital to identify the nodes representing causes and their effects; see e.g.,~\cite{Peters2017}. Beyond causal discovery, there is a need for methodological advances to tackle challenging digraph topology identification problems.  
\subsection{Recovering the Fundamental Structure of Signals}\label{Ss:Learn_Structure}

Consider a network represented as a weighted digraph $\ccalG$, with a node set $\ccalN$ of known cardinality $N$, and an edge set $\ccalE$ of ordered pairs of elements in $\ccalN$. The edge weights $A_{ij}\in\reals$ such that $A_{ij}\neq 0$ for all $(i,j)\in\ccalE$ are collected in the (generally non-symmetric) adjacency matrix $\bbA$. As a more general algebraic descriptor of network structure, one can define a \emph{graph-shift operator} (GSO) $\bbS\in\reals^{N\times N}$ as any matrix having the same sparsity pattern as that of $\ccalG$~\cite{SandryMouraSPG_TSP13}. Accordingly, $\bbS$ can be viewed as a local diffusion operator. 

Our focus in this paper is on identifying digraphs that explain the structure of a random signal.
Formally, let $\bby=[y_1,...,y_N]^\top \in\mbR^N$ be a graph signal in which the $i$th element $y_i$ denotes the signal value at node $i$ of an \emph{unknown digraph} $\ccalG$ with shift operator $\bbS$. Further suppose that we are given a zero-mean signal $\bbx$ with covariance matrix $\bbC_\bbx=\E{\bbx\bbx^\top}$. We say that the sparse digraph $\bbS$ represents the structure of the signal $\bby$ if there exists a diffusion process in the GSO $\bbS$ -- modeled as a polynomial graph filter -- that generates the observation $\bby$ from the input signal $\bbx$ via linear filtering, i.e.,
\begin{equation}\label{eqn_diffusion}
	\textstyle \bby\  =\ \alpha_0 \prod_{l=1}^{\infty} (\bbI-\alpha_l \bbS) \bbx
	\  =\ \big(\sum_{l=0}^{\infty}\beta_l \bbS^l\big)\,\bbx,
\end{equation}
for some set of parameters $\{\alpha_l\}$, or equivalently $\{\beta_l\}$. While $\bbS$ encodes only one-hop interactions, each successive application of the GSO in \eqref{eqn_diffusion} percolates $\bbx$ over $\ccalG$. The justification to say that $\bbS$ represents the structure of $\bby$ is that we can think of the edges of $\bbS$ as direct (one-hop), directional relations between the elements of the signal. The diffusion described by \eqref{eqn_diffusion} generates indirect relations, for instance, longer-range statistical dependencies in the covariance matrix $\bbC_\bby=\E{\bby\bby^\top}$. 
Our goal is to exploit this model to recover the fundamental relations described by $\bbS$ from a set $\ccalY$ of independent realizations of a random signal $\bby$ 
along with prior knowledge of $\bbC_\bbx$. The required statistical information about $\bbx$ is the price paid to accommodate non-stationary processes $\bby$ with respect to (w.r.t.) (possibly) asymmetric GSOs~\cite{marques2016stationaryTSP16,RSSSAMGM_OJSP21}. 

%
\subsection{Paper Outline and Contributions}\label{Ss:Outline}

In Section~\ref{S:prelim_problem}, we formulate the problem of identifying a GSO that explains the fundamental structure of a random signal diffused on a digraph. We advocate an innovative two-step approach whereby: i) given independent output observations and prior information on the input statistics, we identify the graph convolutional filter; and ii) given the filter estimate along with structural constraints on the GSO, we recover the topology of the digraph via convex optimization. 
In Section~\ref{S:EstimatingDirectedFilterFromQuadratic}, we address the problem of identifying the diffusion filter [cf. i)], which entails solving a system of \emph{quadratic} matrix equations formed using the available information on the input along with output signal measurements. 
We show that the set of feasible filters is related to the square roots of the observations' covariance matrix, and that such a set is markedly larger than its counterpart for symmetric filters~\cite{RSSSAMGM_OJSP21}. Conditions under which the system identification task admits a unique solution are derived as well (Section \ref{Ss:perf_obs_feas_set}). Building on these insights, in Section \ref{Ss:imperf_obs_filt_inf} we recast the filter-inference problem as a smooth quadratic minimization subject to Stiefel manifold constraints. Such nonconvex problem can be tackled leveraging recent advances for orthogonality-constrained optimization; see e.g.,~\cite{boumal2023intromanifolds}. 
Subsequent identification of a directed GSO given the diffusion filter estimate [cf. ii)] is addressed in Section \ref{S:EstimatingDirectedShiftFromFilter}. The focus is on finding a \textit{sparse} and structurally admissible shift that commutes with the given filter, thus forcing the latter to be a polynomial in the GSO as in \eqref{eqn_diffusion}. In Section \ref{S:Joint_Estimation} we also propose a \emph{joint} graph filter and topology identification formulation and associated provably convergent algorithm, closing the loop between steps i) and ii) to offer improved sample complexity.
A comprehensive experimental evaluation in Section~\ref{S:Simulations} corroborates the effectiveness of the proposed approach in recovering synthetic graphs in a variety of controlled settings designed to probe different key aspects of the problem. Real data tests hint at the algorithm's  potential impact on urban mobility analyses and portfolio optimization. Concluding remarks with a discussion of limitations along with an outlook towards future work are given in Section \ref{S:Conclusions}.

%
\subsection{Relation to Prior Work}\label{Ss:Prior_Work}

Early topology identification approaches infer (symmetric) graphs whose edge weights correspond to nontrivial correlations between nodal signals~\cite{giannakis18pieee,SI_SPMAG,sihag2025spmag}. To control for spurious correlations due to latent confounders, alternative statistical methods rely on inference of partial correlations~\cite[Ch. 7.3.2]{kolaczyk2009book}. Under Gaussianity assumptions,
this line of work has well-documented connections with sparse precision matrix estimation~\cite{GLasso2008}. Extensions to digraphs include
structural equation models (SEMs)~\cite{BainganaInfoNetworks,shen2017tensors,shen2020jointdiag,han2026covmatching} with acyclicity constraints~\cite{zheng2018dags,saboksayr2023colide}, Granger causality~\cite{Brovelli04Granger}, or
their nonlinear (e.g., kernelized) variants~\cite{giannakis18pieee}. GSP-inspired topology identification frameworks postulate that the network exists as a latent underlying structure, and that observations are generated as a result of a network process defined in such a graph~\cite{DongLaplacianLearning,MeiGraphStructure, Kalofolias2016inference_smoothAISTATS16,segarra2016topoidTSP16,pasdeloup2016inferenceTSIPN16,thanou17,dong_2019_learning,buciulea2022tsipn}. Different from~\cite{DongLaplacianLearning,Kalofolias2016inference_smoothAISTATS16,saboksayr20,saman2021spl}
that infer structure from signals assumed to be smooth over the sought undirected graph, here
the measurements are related to the graph via convolutional filtering. Works that explored this approach recover a symmetric GSO from its eigenvectors, either assuming that the input is white~\cite{segarra2016topoidTSP16,pasdeloup2016inferenceTSIPN16} -- equivalently implying $\bby$ is graph stationary~\cite{marques2016stationaryTSP16}; or, colored as in the present paper~\cite{RSSSAMGM_OJSP21}. Here instead, we distinctly address the general case of digraphs. Relative to~\cite{han2026covmatching,shen2017tensors,shen2020jointdiag} that rely on a specific single-pole graph filter stemming from the SEM model, the filter structure underlying \eqref{eqn_diffusion} can be arbitrary, but the focus here and in~\cite{han2026covmatching} is on learning time-invariant graphs. Although~\cite{shen2020jointdiag} does not require prior knowledge of $\bbC_{\bbx}$, the joint diagonalization method therein assumes that the connectivity patterns of a small subset of nodes are available. 

Relative to the conference precursor~\cite{RSSSAMGM_dsw18}, here we consider digraph topology inference via graph filter identification through a unified presentation along with full technical details (including extended discussions, algorithms, and unpublished theoretical results with their proofs), supported by comprehensive numerical experiments. Noteworthy novel pieces in this full-blown journal paper include: (i) the joint digraph and filter identification formulation; (ii) custom-made algorithms for the open- and closed-loop approaches; (iii) theory supporting the feasibility of the asymmetric diffusion filter identification task; and (iv) a comprehensive performance evaluation protocol.  The latter offers new synthetic test cases, comparisons with baseline methods, as well as real-data studies of Uber commute patterns in New York City and portfolio evaluation.
\vspace{2pt}

\noindent \textbf{Notational conventions.} The entries of a matrix $\mathbf{X}$ and a (column) vector $\mathbf{x}$ are denoted by $X_{ij}$ and $x_i$, respectively. For a three-way tensor we write $\underline{\bbX}$. Sets are represented by calligraphic capital letters.  
The notation $(\cdot)^\top$ stands for matrix transpose, $\mathbf{0}$ and $\mathbf{1}$ refer to the all-zero and all-one vectors; while $\bbI$ denotes the identity matrix. For a vector $\bbx$, $\diag(\mathbf{x})$ is a diagonal matrix whose $i$th diagonal entry is $x_i$. 
For a matrix $\bbX$, $\| \bbX \|_p$ denotes the $\ell_p$ norm of its vectorization $\text{vec}(\bbX)$, while $\| \bbX \|_F$ stands for Frobenius norm. The largest singular value of matrix $\bbX$ is denoted by $\sigma_{\max}(\bbX)$.

\section{Preliminaries and Problem Statement}\label{S:prelim_problem}

Suppose we observe realizations of a random signal $\bby$ generated through diffusion on $\ccalG$ of an input $\bbx$, namely via successive applications of a GSO $\bbS$ as in \eqref{eqn_diffusion}. Since the (statistical) properties of the signal $\bby$ depend on $\bbS$, our goal is to use a set of observations together with available information on the excitation input to infer the digraph topology. In other words, we aim to recover the GSO which encodes pairwise influence between graph nodes, given observable indirect relationships generated by a diffusion process.  
To formally state the problem, we elaborate on the diffusion model in \eqref{eqn_diffusion} as well as on the available information from the graph signals. 

%
\subsection{Graph Filtering as a Model of Network Diffusion}\label{Ss:Diffusion}

While the diffusion expressions in \eqref{eqn_diffusion} entail (possibly) infinite-degree polynomials in $\bbS\in\reals^{N\times N}$, the Cayley-Hamilton theorem asserts that they are equivalent to polynomials of degree smaller than $N$. Upon defining the vector of coefficients $\bbh:=[h_0,\ldots,h_{L-1}]^\top$ and the asymmetric graph filter $\bbH:=\sum_{l=0}^{L-1} h_l \bbS^l$~\cite{SandryMouraSPG_TSP13}, the model in \eqref{eqn_diffusion} becomes
\begin{align}\label{E:Filter_input_output_time}
	\bby  = \textstyle \big(\sum_{l=0}^{L-1}h_l \bbS^l\big)\,\bbx
	= \bbH \bbx
\end{align}
for some $\bbh$ and $L\leq N$. Most germane to the present paper is that since $\bbH$ is a polynomial in $\bbS$, 
then $\bbH$ and $\bbS$ commute, i.e., $\bbH \bbS = \bbS \bbH$. We will exploit this identity in Sections \ref{S:EstimatingDirectedShiftFromFilter} and \ref{S:Joint_Estimation}, which can be interpreted to imply that graph filters are shift invariant. Another upshot of the polynomial relationship is that the eigenvectors of $\bbH$ and $\bbS$ coincide, provided $\bbS$ is diagonalizable which we henceforth assume. 
Hence, while the diffusion implicit in $\bbH$ obscures part of the structure of $\bbS$ (its eigenvalues), its eigenvectors remain as templates of the underlying network topology~\cite{pasdeloup2016inferenceTSIPN16,segarra2016topoidTSP16,RSSSAMGM_OJSP21}.

%
\subsection{Observations and Prior Information on the Input Signals}\label{Ss:Prior_Information}

We observe $M$ network processes $\{\bby_m\}_{m=1}^{M}$ on $\ccalG$, each one corresponding to a different input zero-mean random signal $\bbx_m$ that is diffused via a common filter $\bbH$. The multiplicity and statistical diversity of input processes will be instrumental towards identifying (uniquely) the diffusion filter. Let $\ccalY_m\!:=\!\{\bby_m^{(p)}\}_{p=1}^{P_m}$ capture the observed output signal realizations associated with the $m$th process, and likewise let $\ccalY \! := \! \bigcup_{m=1}^M \ccalY_m$ collect all available observations. Regarding the $m$th input process, one could conceivably assume its covariance matrix $\bbC_{\bbx,m}$, or even realizations of the signal $\{\bbx_m^{(p)}\}_{p=1}^{P_m}$ are given. Henceforth, the focus will be on the most pragmatic setting whereby only second-order statistics are available~\cite{RSSSAMGM_OJSP21,shen2017tensors,han2026covmatching}. 
A motivating example can be drawn from opinion formation models in social networks capturing influence among agents. Suppose (e.g., socio-economic or demographic) features for each of the agents are measurable, while initial beliefs on different topics $m=1,\ldots,M$ (namely the entries of $\bbx_m$) are challenging to acquire. It is conceivable that agents with similar features will exhibit correlated beliefs (in a way akin to homophily), hence one could build a model of $\bbC_{\bbx,m}$ based on said features. As a second example, consider financial networks comprising stocks as nodes and their interdependencies as directed links. Publicly-traded stock prices $(\bby_m)$ are known to depend on stock purchases by investors $(\bbx_m)$, whose details are
often hidden from the public for privacy and strategic reasons. On the other hand, each publicly-traded company may broadcast e.g., quarterly statistical summaries of purchases of its stock $(\bbC_{\bbx,m},\:m=1,\ldots,4)$; see for example the setting considered in~\cite{shen2017tensors} and the portfolio optimization test case in Section \ref{S:Simulations}.

%
\subsection{Problem Statement}\label{Ss:Prob_Statement}

Given observations $\ccalY \! = \! \bigcup_{m=1}^M \ccalY_m $ adhering to the generative model \eqref{E:Filter_input_output_time}, where a common filter $\bbH:=\sum_{l=0}^{L-1} h_l \bbS^l$ diffuses $M$ zero-mean inputs $\bbx_m$ with known covariance matrices $\bbC_{\bbx,m}=\E{\bbx_m\bbx_m^\top}$, $m=1,\ldots,M$, the goal is to find the sparsest shift $\bbS$ with desirable topological properties (e.g., it is a valid adjacency matrix) that is consistent with the observations. The order $L\leq N$ of the convolutional graph filter $\bbH$ and its polynomial coefficients $\bbh\in\reals^{L}$ are unknown. We also do not assume a specific structure, say $\bbH=(\bbI-\bbS)^{-1}$ induced by a linear SEM~\cite{shen2017tensors,shen2020jointdiag,han2026covmatching}. 

\begin{figure}[t]
	\begin{minipage}[b]{.48\linewidth}
		\centering
		\includegraphics[width=\linewidth, trim={0cm, 0cm, 0cm, 1cm}]{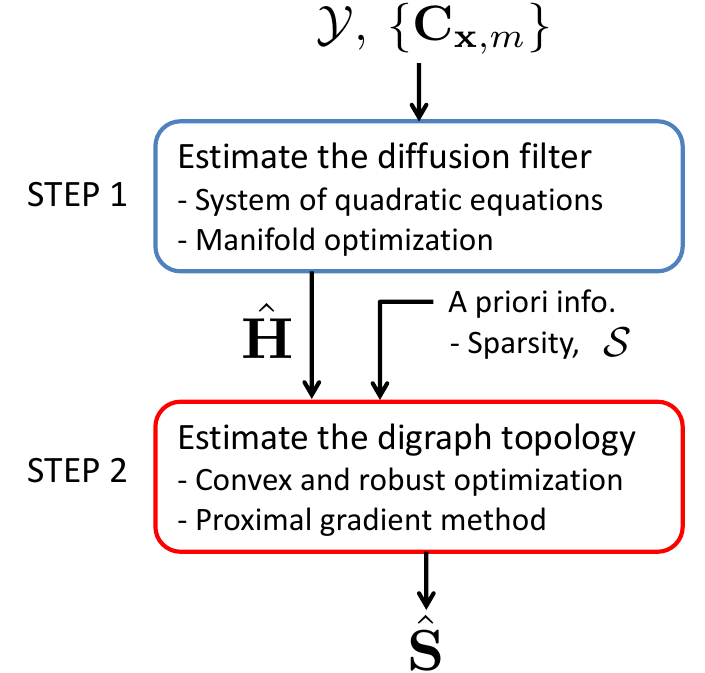}
	\end{minipage}
	%
	%
	\begin{minipage}[b]{.48\linewidth}
		\centering
		\includegraphics[scale=0.355, trim={0cm, 0cm, 0cm, 1cm}]{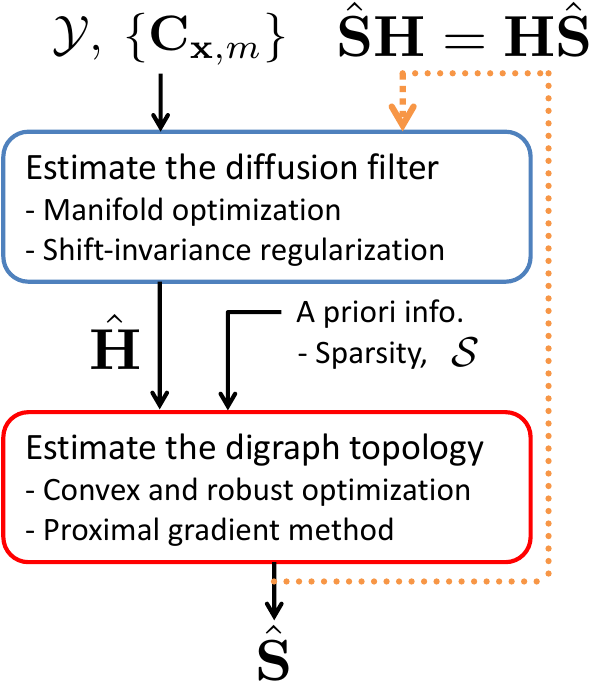}
	\end{minipage}
	%
	\caption{{Schematic view of the digraph identification methods in their (left) open loop and (right) closed loop renditions.}} 
	\label{F:scheme_topo_id}\vspace{-10pt}
\end{figure}

Motivated by the discussion following \eqref{E:Filter_input_output_time}, our two-step network topology inference approach is to: i) first use realizations of observed signals in $\ccalY$ together with side information on the excitation inputs $\{\bbx_m\}_{m=1}^M$ to \textit{identify the diffusion filter} $\bbH$; and ii) then use the estimated filter along with prior information on the network topology to \textit{infer the GSO} $\bbS$.  Step i) is addressed in Section \ref{S:EstimatingDirectedFilterFromQuadratic}, while ii) is discussed in Section \ref{S:EstimatingDirectedShiftFromFilter}. A joint graph filter and topology identification algorithm is developed in Section \ref{S:Joint_Estimation}, which alternates between steps i) and ii) in a mutually reinforcing fashion. Refer to Fig. \ref{F:scheme_topo_id} for a schematic depiction of both approaches.

%

\section{Asymmetric diffusion filter identification}\label{S:EstimatingDirectedFilterFromQuadratic}

Here we study the system identification step, a challenging problem that is of independent interest beyond our topology identification end goal. In a number of applications, realizations of the excitation input $\bbx_m$ may be challenging to acquire, but information about the \textit{statistical} description of $\bbx_m$ could still be available. As in our statement of the problem, assume 
that the excitation inputs are zero mean and their covariance matrices $\bbC_{\bbx,m}$ are known. As explained in Section \ref{S:prelim_problem}, suppose also that for each $m=1,\ldots,M$ we acquire observations $\ccalY_m=$ $\{\bby_m^{(p)}\}_{p=1}^{P_m}$, which are then used to estimate the output covariance $\bbC_{\bby,m} = \EE [\bby_m\bby_m^\top]$ via sample averaging, that is
\begin{equation}\label{E:cov_estimate}
	\hbC_{\bby,m}=\frac{1}{P_m-1}\sum_{p=1}^{P_m}\bby_m^{(p)}(\bby_m^{(p)})^\top.
\end{equation}
Since under \eqref{E:Filter_input_output_time} the output covariance is given by $\bbC_{\bby,m} = \bbH \bbC_{\bbx,m}\bbH^\top$, the aim is to identify a filter $\bbH$ such that $\hbC_{\bby,m}$ and $\bbH \bbC_{\bbx,m} \bbH^\top$ are close in some sense. This ``covariance matching'' idea was first proposed  in~\cite{shen2016tensors_SEM_Asil16} for single-pole SEM filters and subsequently generalized for general filters in~\cite{RSSSAMGM_icassp17,RSSSAMGM_dsw18,RSSSAMGM_OJSP21}; see~\cite{han2026covmatching} for a fresh look at the linear SEM case.


\subsection{Solving Quadratic Equations for Filter Identification}\label{Ss:perf_obs_feas_set}

Assuming for now perfect knowledge of the output signal covariances, the above rationale suggests studying the solutions of the following system of matrix quadratic equations
\begin{equation}\label{E:quadratic_system}
	\bbC_{\bby,m}=\bbH \bbC_{\bbx,m} \bbH^\top, \quad m=1,\ldots,M,
\end{equation}
for real-valued and possibly asymmetric $\bbH$. To study the solutions of \eqref{E:quadratic_system}, first consider $M=1$ and henceforth drop the subindex $m$ to focus on the square roots of $\bbC_{\bby}=\bbH \bbC_{\bbx} \bbH^\top$. 

To that end, recall first that the principal square root of $\bbC_{\bby}$ is the only symmetric and positive semidefinite (PSD) matrix $\bbC_{\bby}^{1/2}$ which satisfies $\bbC_{\bby}=\bbC_{\bby}^{1/2}\bbC_{\bby}^{1/2}$. Such a matrix is given by $\bbC_{\bby}^{1/2}=\bbV_{\bby}\bbLambda_{\bby}^{1/2}\bbV_{\bby}^\top$, where $\bbV_{\bby}$ are the eigenvectors of $\bbC_{\bby}$ and $\bbLambda_{\bby}^{1/2}$ stands for a diagonal matrix with the square roots of the (non-negative) eigenvalues of $\bbC_{\bby}$. 
With $\bbU$ denoting an \textit{orthogonal} matrix (such that $\bbU\bbU^\top=\bbU^\top\bbU=\bbI$), one can show that any square matrix factor $\bbH_\bbx$ such that $\bbC_\bby=\bbH_\bbx \bbH_\bbx^\top$ is of the form $\bbH_{\bbx}=\bbC_{\bby}^{1/2}\bbU$. This observation can be leveraged to establish the following result, which characterizes the solution set of each individual quadratic equation in \eqref{E:quadratic_system}.

\begin{myproposition}\label{P:sol_quadratic_system}
If $\bbC_{\bbx,m}$ and $\bbC_{\bby,m}$ are of full rank, the set $\ccalH_m$ containing all the (possibly asymmetric) matrices $\bbH$ that solve \eqref{E:quadratic_system} for a particular $m$ is given by 
	\begin{equation}
		\!\!\!\!\ccalH_m=\left\{\bbH\;\given \;\bbH \!=\! \bbC_{\bby,m}^{1/2} \bbU \bbC_{\bbx,m}^{-1/2}\;\;\text{and}\;\;\bbU\bbU^\top=\bbI\right \}.\label{E:feas_H_non_sym_v3}
	\end{equation}
\end{myproposition}
\begin{myproof}A simple substitution suffices to show that every $\bbH$ of the form in \eqref{E:feas_H_non_sym_v3} solves the $m$th equation in \eqref{E:quadratic_system}. Conversely, given an $\bbH$ that solves \eqref{E:quadratic_system}, form the matrix $\bbU = \bbC_{\bby,m}^{-1/2} \bbH \bbC_{\bbx,m}^{1/2}$ and  observe that if $\bbU$ is orthogonal, then $\bbH \in \ccalH_m$. 
Orthogonality of $\bbU$ follows since $\bbU \bbU^\top = \bbC_{\bby,m}^{-1/2} \bbH \bbC_{\bbx,m} \bbH^\top \bbC_{\bby,m}^{-1/2} = \bbI$, where the last equality comes from the fact that $\bbH$ is a solution of \eqref{E:quadratic_system}. 
\end{myproof}

As expected, the solution to each of the matrix quadratic equations \eqref{E:quadratic_system} is not unique. In fact, Proposition \ref{P:sol_quadratic_system} asserts that $\ccalH_m$ can be parameterized by means of the orthogonal matrix $\bbU$. If the GSO $\bbS$ (and hence the filter $\bbH$) is symmetric, then $\bbU=\diag(\bbb)$, with $\bbb=\{-1,1\}^N$. 
Symmetry removes the rotational ambiguity but $2^N$ sign permutations remain. This added structure reduces considerably the size of the feasible set in \eqref{E:feas_H_non_sym_v3}; see \cite{RSSSAMGM_OJSP21} for details. Still, even in this case it is apparent that $\bbH$ is non-identifiable for $M=1$.

For the general case of $M>1$, the set of solutions to the system of quadratic equations \eqref{E:quadratic_system} is given by the intersection of \eqref{E:feas_H_non_sym_v3} for all diffusion processes, i.e.,  $\ccalH_{1:M} \!:= \! \bigcap_{m=1}^M\ccalH_m$.
In the next lemma, we show how the feasible region $\ccalH_{1:M}$ shrinks for $M>1$ and under mild pragmatic assumptions.

\begin{mylemma}\label{L:identifiability_directed}
Consider the system of equations \eqref{E:quadratic_system} for $M>1$ and assume $\{\bbC_{\bbx,m}\}_{m=1}^{M}$ and $\{\bbC_{\bby,m}\}_{m=1}^{M}$ are of full rank. 
	Define the matrices $\{\bbC_{\bby\bby,m}\}_{m=1}^{M}$ and $\{\bbC_{\bbx\bbx,m}\}_{m=1}^{M}$ such that
\begin{equation}
\begin{aligned} \label{e:C_yyy}
\bbC_{\bby\bby,m} & := \bbC_{\bby,m}^{1/2} \bbC_{\bby,m\!-\!1}^{-1} \bbC_{\bby,m}^{1/2}, \\
\bbC_{\bbx\bbx,m} & := \bbC_{\bbx,m}^{1/2} \bbC_{\bbx,m\!-\!1}^{-1} \bbC_{\bbx,m}^{1/2},
\end{aligned}
\end{equation}
for all $m=1, \ldots, M$ where, for notational consistency, we define $\bbC_{\bby,0} = \bbC_{\bby,M}$ and $\bbC_{\bbx,0} = \bbC_{\bbx,M}$.
If $\{\bbC_{\bbx\bbx,m}\}_{m=1}^M$ have distinct eigenvalues, then the set of solutions of \eqref{E:quadratic_system} are
	\begin{align}\label{E:solution_set_asym}
	\ccalH_{1:M}\!\!=&\bigcap_{m=1}^M\left\{\bbH\;\given \;\bbb_m\!\in\!\{-1,1\}^N\;\;\text{and}\;\;\right.\nonumber\\
	&\left.\bbH\!=\!\bbC_{\bby,m}^{1/2}\bbV_{\mathbf{yy},m}\diag(\bbb_m)\bbV_{\mathbf{xx},m}^\top\bbC_{\bbx,m}^{-1/2}\right\},
	\end{align}
	where matrices $\{\bbV_{\bby\bby,m}\}_{m=1}^{M}$ and $\{\bbV_{\bbx\bbx,m}\}_{m=1}^{M}$ are the eigenvectors of $\{\bbC_{\bby\bby,m}\}_{m=1}^{M}$ and $\{\bbC_{\bbx\bbx,m}\}_{m=1}^{M}$, respectively.
\end{mylemma}

\begin{myproof}
See Appendix \ref{L:identifiability_directed_proof}.
\end{myproof}

Inspection of \eqref{E:solution_set_asym} reveals that $M\!=\!2$ processes shrink the (continuous) manifold-constrained feasible set to $2^N$ possible alternatives. Intuitively, each $m$ provides a new set of observations that reduces the original $N^2$ degrees of freedom in $\bbH$ to $2^N$ discrete choices.  It is thus conceivable that, as $M$ grows, the cardinality of $\ccalH_{1:M}$ shrinks and the problem is rendered identifiable (up to any unavoidable sign ambiguity inherent to any quadratic equation). In this context, we show that when $M=4$, even if the covariances $\{\bbC_{\bbx,m}\}_{m=1}^{4}$ of the inputs partially share eigenvectors, uniqueness can be established as long as their eigenvalues are sufficiently different.

\begin{mytheorem}\label{T:new_uniqueness}
	Consider the system of quadratic equations~\eqref{E:quadratic_system} for $M=4$ and suppose that $\bbC_{\bbx,m} = \bbV_{\bbx} \diag (\bblambda_m) \bbV_{\bbx}^\top$ for $m=1, 2$ and $\bbC_{\bbx,m} = \bbW_{\bbx} \diag (\bblambda_m) \bbW_{\bbx}^\top$  for $m = 3,4$. 
	Then,~\eqref{E:quadratic_system} has a unique solution, i.e., the filter $\bbH$ is identifiable up to a sign ambiguity, if the following conditions hold:\vspace{2pt}
    
	\noindent i) All eigenvalues in $\bblambda_m$ are distinct for all $m$;\vspace{2pt}
    
	\noindent ii) $\lambda_{1,i} \lambda_{2,j} \neq \lambda_{1,j} \lambda_{2,i}$ and $\lambda_{3,i} \lambda_{4,j} \neq \lambda_{3,j} \lambda_{4,i}$ for all $i \neq j$;\vspace{2pt}
    
	\noindent iii) The space spanned by any $k$ eigenvectors in $\bbV_{\bbx}$ is different from that spanned by any $k$ eigenvectors in $\bbW_{\bbx}$, for $k=1, \ldots, N-1$; and\vspace{2pt}
    
	\noindent iv) $\mathrm{rank}(\bbH) = N$.
\end{mytheorem}

\begin{myproof}
See Appendix \ref{T:new_uniqueness_proof}.
\end{myproof}

Theorem~\ref{T:new_uniqueness} gives sufficient conditions under which the asymmetric filter $\bbH$ is identifiable from the second-order statistics of only $M=4$ diffusion processes, arranged in two pairs that share the input eigenbases $\bbV_{\bbx}$ ($m=1,2$) and $\bbW_{\bbx}$ ($m=3,4$). It complements Lemma~\ref{L:identifiability_directed} by handling a structured case in which inputs partially share eigenvectors, trading spectral diversity \emph{within} each pair for diversity \emph{across} the two input subspaces. Conditions i) and ii) act on the input eigenvalues: i) ensures each input covariance has a well-defined eigenbasis, while the Kruskal-rank condition ii) secures the essential uniqueness of the PARAFAC factorizations used in the proof, pinning down $\bbH$ up to the $2^N$ sign patterns already exposed by Lemma~\ref{L:identifiability_directed}. Condition iii) removes this residual ambiguity by forcing $\bbV_{\bbx}$ and $\bbW_{\bbx}$ to be sufficiently different, and iv) requires the filter to be invertible. Taken together, the conditions formalize a simple intuition: sufficient statistical diversity of the excitation, both within and across the two pairs, is what disambiguates an asymmetric filter that a single process leaves undetermined up to a rotation.

\subsection{Graph Filter Identification Algorithm}\label{Ss:imperf_obs_filt_inf}

In practice, only empirical covariances \eqref{E:cov_estimate} are available and the equalities in \eqref{E:quadratic_system} must be relaxed.  Given estimates $\{\hbC_{\bby,m}\}_{m=1}^M$ obtained with sufficient samples to ensure full-rankness, our digraph filter identification approach is to solve the manifold-constrained least-squares (LS) problem [cf. \eqref{E:feas_H_non_sym_v3}]
\begin{align}\label{E:general_problem_non_symmetric_shifts} 
	\min_{ \bbH,\{\bbU_m\}} \; &\frac{1}{2}\sum_{m=1}^{M}\| \bbH-\hbC_{\bby,m}^{1/2} \bbU_m \bbC_{\bbx,m}^{-1/2} \|_F^2  \\
	\text{s. to }\;
	&\bbU_m\in\ccalU_N,\,\:\:m=1,\ldots,M, \nonumber
\end{align}
where $\ccalU_N:=\{\bbU\in\reals^{N\times N}\:\given\: \bbU^\top\bbU=\bbU\bbU^\top=\bbI\}$ denotes the Stiefel manifold of $N\times N$ real orthogonal matrices.  If the accuracy of the empirical covariances $\hbC_{\bby,m}$ differs significantly across diffusion processes $m=1,\ldots, M$, it may be prudent to introduce non-uniform coefficients to weigh the residuals taking into account those accuracies (possibly as function of the sample sizes $P_m$). Indeed, the residuals in \eqref{E:general_problem_non_symmetric_shifts} should vanish in a noiseless setting so the criterion minimizes this discrepancy across the $M$ processes considered. 

Although the smooth objective
\begin{equation}\label{e:joint_objective} 
\phi(\bbH,\{\bbU_m\}) \: :=\: \frac{1}{2}\sum_{m=1}^M 
\| \bbH - \hbC_{\bby,m}^{1/2} \bbU_m \bbC_{\bbx,m}^{-1/2}\|_F^2 
\end{equation}
in \eqref{E:general_problem_non_symmetric_shifts} is jointly convex in the unknowns $\bbH$ and $\{\bbU_m\}_{m=1}^M$, the constraint set $\{\bbU_m\}_{m=1}^M\in\ccalU_N^M$ is not (here $\ccalU_N^M:=\ccalU_N\times\ldots\times\ccalU_N$ denotes the product manifold). We solve \eqref{E:general_problem_non_symmetric_shifts} via an inexact alternating minimization (IAM) algorithm on $\ccalU_N^M$.
Specifically, at each (outer) iteration $k=1,2,\ldots$, given $\bbH^k$ the problem \eqref{E:general_problem_non_symmetric_shifts} is separable w.r.t. to each of the $\{\bbU_m\}_{m=1}^M$. Hence, these subproblems can be tackled concurrently, and we solve them inexactly by running a few (inner) iterations of Riemannian gradient descent (GD)~\cite{boumal2023intromanifolds}, while $\bbH^k$ is fixed to its most up-to-date value and subsequently updated in closed form to complete the cyclic update.

In more detail, the update of $\bbU_m$ at iteration $k$ amounts to running $t=0,1,\ldots,t_{\max}$ inner iterations of Riemannian GD on $\bbU^t$ initialized as $\bbU^{0}=\bbU_m^k$. To this end, one first computes the Riemannian gradient of $\phi(\bbH,\{\bbU_m\})$ w.r.t. $\bbU_m$, denoted by $\grad_{\bbU_m} \phi$. Letting $\bbG_m^t := \nabla_{\bbU_m} \phi(\{\bbU_m\},\bbH^k)\given_{\bbU_m=\bbU^t}$ be the Euclidean gradient, $\grad_{\bbU_m} \phi(\bbU^t)$ is the projection of  
\begin{equation}\label{e:nabla_fm}
\bbG_m^t = \hbC_{\bby,m}^{1/2} \big( \hbC_{\bby,m}^{1/2} \bbU^t \bbC_{\bbx,m}^{-1/2}-\bbH^k\big)  \bbC_{\bbx,m}^{-1/2}
\end{equation}
onto the tangent space $\mathrm{T}_{\bbU^t}\ccalU_N$~\cite[pp. 161-162]{boumal2023intromanifolds}. This yields
\begin{equation}\label{e:grad_fm}
\grad_{\bbU_m} \phi(\bbU^t) = \bbG_m^t-\bbU^t\frac{(\bbU^t)^\top \bbG_m^t +(\bbG_m^t)^\top\bbU^t}{2}.
\end{equation}
Having computed the gradient, a classical descent method consists of taking a certain step in the opposite direction. However, this step likely results in a point outside of the manifold, so we have to project it back to $\ccalU_N$. This projection might be computationally intensive, so the \emph{retraction} alternative is used instead. All in all, the Riemannian GD step is given by 
\begin{equation}\label{eq:r_grad_step}
\bbU^{t+1} = \textrm{Retr}_{\bbU^t}(-\alpha\grad_{\bbU_m} \phi(\bbU^t)),
\end{equation}
where we adopt the polar retraction $\textrm{Retr}_{\bbU}(\bbW)=\bbP\bbQ^\top$~\cite[p. 161]{boumal2023intromanifolds}, with $\bbP\bbSigma\bbQ^\top=\textrm{svd}[\bbU+\bbW]$. Step size $\alpha>0$ is chosen via backtracking line search, e.g., using the Armijo rule. A constant step size can also be used, informed by the Lipschitz constant of $\grad_{\bbU_m} \phi$. After completing $t_{\max}$ gradient updates as in \eqref{eq:r_grad_step}, the $m$th inner loop returns $\bbU_m^{k+1}=\bbU^{t_{\max}}$.

To conclude the cycle in outer iteration $k$, we update $\bbH^k$. Since \eqref{E:general_problem_non_symmetric_shifts} is an unconstrained quadratic problem w.r.t. $\bbH$, we can update it in closed form as
\begin{equation}\label{E:filter_estimate}
	\bbH^{k+1}=\frac{1}{M}\sum_{m=1}^M\hbC_{\bby,m}^{1/2} \bbU_m^{k+1} \bbC_{\bbx,m}^{-1/2}.
\end{equation}
The overall IAM procedure for asymmetric graph filter identification is tabulated under Algorithm \ref{alg:topoid_directed}.\vspace{2pt}

\begin{algorithm}[t]
	\caption{IAM for asymmetric graph filter identification}
	\label{alg:topoid_directed}
	\begin{algorithmic}[1]
		\REQUIRE $\{\bbC_{\bbx,m}\}_{m=1}^{M}$, $\{\hbC_{\bby,m}\}_{m=1}^{M}$, parameters $t_{\max},\epsilon> 0$.
		\STATE \textbf{initialize} $k=0$, $\{\bbU_m^0\}_{m=1}^{M} \in\ccalU_N$ and $\bbH^0$ at random.
		\REPEAT
		\FOR{$m=1,\ldots,M$ in parallel}
		\STATE  \textbf{initialize} $t=0$ and $\bbU^0 := \bbU_m^k$.
		\REPEAT
		\STATE Compute $\bbG_{m}^{t}$ using \eqref{e:nabla_fm}.
		\STATE Compute $\grad_{\bbU_m}\phi(\bbU^t)$ using \eqref{e:grad_fm}.
        \STATE Choose step size $\alpha$ via the Armijo rule.
        \STATE Update $\bbU^{t+1}$ via Riemannian GD step in \eqref{eq:r_grad_step}.
		\STATE  $t \gets t+1$.
		\UNTIL{$t=t_{\max}$}.
		\RETURN $\bbU_m^{k+1} := \bbU^t$.
		\ENDFOR
		\STATE Update $\bbH^{k+1}$ using \eqref{E:filter_estimate}.
		\STATE $k \gets k+1$.
		\UNTIL{$\| \bbH^{k} - \bbH^{k-1}  \|_{F} / \| \bbH^{k-1} \|_{F} \leq \epsilon$}.
		\RETURN Graph filter estimate $\hbH:=\bbH^{k}$.
	\end{algorithmic}
\end{algorithm}

\noindent\textbf{Complexity and convergence.} Algorithm~\ref{alg:topoid_directed} provably converges to a stationary point of the nonconvex problem \eqref{E:general_problem_non_symmetric_shifts}; see e.g.,~\cite{peng2023bcdmanifold} and our more general result in Theorem \ref{T:alg_convergence}. While global optimality guarantees are elusive, numerical tests in Section~\ref{S:Simulations} corroborate the effectiveness of the proposed optimization strategy. The computational complexity of Algorithm~\ref{alg:topoid_directed} is $\mathcal{O}(N^3)$ per iteration due to the various $N\times N$ matrix multiplications required to compute the gradients, and the SVD involved in the polar retraction. 

\section{Digraph topology inference}\label{S:EstimatingDirectedShiftFromFilter}

Given the graph filter $\bbH$, our approach to infer the topology of the underlying digraph is to find a GSO $\bbS$ that satisfies certain desirable structural properties and is compatible with the convolutional nature of $\bbH$. Focusing on recovery of the sparsest shift operator (i.e., the graph that minimizes the number of direct interactions among nodes), one can solve
\begin{equation}\label{E:general_problem}
	\hat{\bbS} := \argmin_{\bbS \in \ccalS} \
	\|\bbS\|_0 ,   \quad                       
	\text{s.~to }\:\bbH \bbS = \bbS \bbH,
\end{equation}
where $\|\bbS\|_0$ counts the number of non-zero entries of $\bbS$, $\ccalS$ is a convex set specifying the type of GSO we want to identify \cite{segarra2016topoidTSP16}, and the constraint $\bbH \bbS = \bbS \bbH$ forces the filter $\bbH$ to be a polynomial in $\bbS$ (when all the  eigenvalues of $\bbS$ are simple); see \cite{MeiGraphStructure,rey2023tsp} and the discussion following \eqref{E:Filter_input_output_time}. 
This last constraint offers an important departure from the (undirected) graph learning algorithms in~\cite{segarra2016topoidTSP16, pasdeloup2016inferenceTSIPN16, RSSSAMGM_OJSP21}. These approaches first estimate the \textit{eigenvectors} of $\bbH$, and then constrain $\bbS$ to be diagonalized by those eigenvectors in a convex problem to recover the unknown eigenvalues. While in \cite{segarra2016topoidTSP16, pasdeloup2016inferenceTSIPN16, RSSSAMGM_OJSP21} one searches over a lower-dimensional space ($N$ versus $N^2$ here), the formulation \eqref{E:general_problem} avoids computing an eigendecomposition and, more importantly, solving a problem over complex-valued variables. 
For undirected graphs, recent approaches impose the constraint $\bbC_{\bby} \bbS = \bbS \bbC_{\bby}$~\cite{rasoul2020algorithms,buciulea2022tsipn,navarro2022jmlr,liu2023neurips}, a direct consequence of graph stationarity and therefore is not applicable here.

The constraint $\bbS \in \ccalS$ in \eqref{E:general_problem} incorporates a priori knowledge about $\bbS$. If we let $\bbS = \bbA$ for a digraph with non-negative weights and no self-loops, we can impose $\ccalS \!:= \! \{ \bbS \, | \, S_{ij} \geq 0, \;\,   S_{ii} = 0, \;\, \textstyle\sum_i S_{i1} \! = \! 1 \}$. The last condition fixes the scale of the admissible graphs by setting the weighted in-degree of the first node to $1$, and rules out the trivial solution $\bbS\!=\!\bbzero$. 

\subsection{Convex and Robust Formulation}\label{Ss:Convex_Robust}

The formulation \eqref{E:general_problem} should be modified to account for model mismatches and imperfect estimates $\hbH$ due to finite-sample effects or local optimal solutions obtained from Algorithm \ref{alg:topoid_directed}. We thus relax the linear matrix equality constraint in \eqref{E:general_problem}, with some measure of the residual $\hbH \bbS - \bbS \hbH$. 
For the numerical tests in Section \ref{S:Simulations}, we adopt a Frobenius-norm error measure and solve the convex $\ell_1$-norm minimization problem
\begin{equation}\label{E:robust_general_problem}
	\hat{\bbS} \in \argmin_{\bbS \in \ccalS} \
	\|\bbS\|_1 ,   \quad                       
	\text{s.~to }\:
	\| \hbH \bbS - \bbS \hbH \|_{F}^2 \leq \varepsilon.
\end{equation}
%
The bound $\varepsilon$ can be selected based on a priori knowledge on the effectiveness of the prior filter identification step.


\begin{algorithm}[t!]
	\caption{PG for digraph topology identification}
	\label{A:alg1}
	\algsetup{linenosize=\normalsize}
	\begin{algorithmic}[1]
		\REQUIRE  $\hbH$,  $\mu>0$. Set $\gamma=1/[4\mu\sigma_{\max}^2(\hbH)]$. 
		\STATE Initialize $k=0$ and $\bbS^0\neq \mathbf{0}$ as a sparse, random matrix. 
		\WHILE {not converged} 
		\STATE Form $\nabla g(\bbS^k)$ using \eqref{eq:grad_g}. 
		\STATE Take GD step $\bbD^k=\bbS^k-\gamma \nabla g(\bbS^k)$. 
		\STATE Update $\bbS^{k+1} = \text{prox}_{\gamma \| \cdot \|_1,\ccalS}\left(\bbD^k \right)$ via the operator \eqref{eq:prox_S_P}. 
		\STATE $k = k + 1$. 
		\ENDWHILE 
		\RETURN $\hbS = \bbS^k$.
	\end{algorithmic}
\end{algorithm}

\subsection{Proximal Gradient Algorithm for Topology Identification}\label{Ss:Batch_Proximal}

Exploiting the problem structure in \eqref{E:robust_general_problem}, we develop a proximal gradient (PG) algorithm to recover the directed network topology from the graph filter estimate $\hbH$; see~\cite{rasoul2020algorithms}. To this end, we dualize $\| \hbH \bbS - \bbS \hbH \|_{F}^2 \leq \varepsilon$ in \eqref{E:robust_general_problem} and write the composite, nonsmooth convex optimization
\begin{equation}\label{eq:opt_batch} 
	\hbS \in \underset{\bbS \in \ccalS}{\text{argmin}} \:
	\varphi(\bbS):= \|\bbS\|_1\!+\!\underbrace{\frac{\mu}{2} \| \hbH \bbS - \bbS \hbH \|_{F}^2}_{g(\bbS)}.
\end{equation}
The function $g(\cdot)$ is convex and $L_g$-smooth [i.e., $\nabla g(\cdot)$ is $L_g$-Lipschitz continuous], with $\mu>0$ a tuning parameter. 

To derive the PG iterations, notice that the gradient of $g(\bbS)$ 
\begin{equation} \label{eq:grad_g}
	\nabla g(\bbS) = \mu\left[\hbH^\top(\hbH \bbS - \bbS \hbH) - (\hbH \bbS - \bbS \hbH)\hbH^\top\right]
\end{equation}
is Lipschitz with constant $L_g\!=\!4 \mu \sigma_{\max}^{2} (\hbH)$.  With $\gamma>0$ and $\ccalS$ a convex set, introduce the proximal operator of a function $\gamma f(\cdot):\reals^{N\times N}\mapsto \reals$ at matrix $\bbD \in \reals^{N \times N}$ as 
\begin{equation*} 
\bbZ(\bbD)=\text{prox}_{\gamma f,\ccalS}(\bbD) := \argmin_{\bbX \in \ccalS} \left[f(\bbX) + \frac{1}{2\gamma} \| \bbX -\bbD \|_F^2\right].
\end{equation*}
Using both ingredients, the PG updates with fixed step size $\gamma < \frac{2}{L_g}$ to solve the digraph topology identification problem \eqref{eq:opt_batch} are given by (once more, $k=1,2,\ldots$ denote iterations)
\begin{equation} \label{eq:batch_prox}
	\bbS^{k+1} := \text{prox}_{\gamma \| \cdot \|_1,\ccalS}\left(\bbS^k - \gamma \nabla g(\bbS^k) \right).
\end{equation}
%

Evaluating the proximal operator efficiently is key to the success of PG methods. For $\ccalS \!:= \! \{ \bbS \, | \, S_{ij} \geq 0, \;\,   S_{ii} = 0, \;\, \textstyle\sum_i S_{i1} \! = \! 1 \}$ and $f(\bbS) = \|\bbS\|_1$, it follows that $\text{prox}_{\gamma \| \cdot \|_1,\ccalS}(\bbD)\in\reals^{N\times N}$ has entries given by
\begin{equation} \label{eq:prox_S_P} 
	\hspace{-0.16cm}[\bbZ(\bbD)]_{ij}=\left\{\begin{array}{lc}
		0, & i=j \\
		\bigtriangleup_{i-1}([D_{21},\ldots,D_{N1}]), & i=2,\ldots,N,\: j=1 \\
		\max(0, D_{ij} - \gamma), & \text{otherwise}
	\end{array}\right.
\end{equation}
where $\bigtriangleup(\cdot): \reals^{N\!-\!1} \mapsto [0,1]^{N\!-\!1}$ is a projection onto the $N-1$ dimensional probability simplex enforcing $\sum_{i} S_{i1}\! =\! 1$. Since $\bigtriangleup$ is an $N-1$ dimensional vector, the notation $\bigtriangleup_i$ used in \eqref{eq:prox_S_P} refers to the $i$th entry of said vector. The projection can be computed in closed form using the method in~\cite[Alg.~1]{chen2011projection}. Beyond the first column of its output, the proximal map nulls the diagonal entries of $\bbS^{k+1}$ and applies a non-negative soft-thresholding operator to update the remaining entries. The PG iterations in \eqref{eq:batch_prox} are tabulated under Algorithm \ref{A:alg1}, which will also serve as a key building block for the \emph{joint} digraph and filter identification algorithm in Section \ref{S:Joint_Estimation}.\vspace{2pt} 

\noindent \textbf{Complexity and convergence.} Just like Algorithm \ref{alg:topoid_directed}, the computational complexity is dominated by the gradient evaluation in \eqref{eq:grad_g}, incurring a cost of $\ccalO(N^3)$ 
per iteration. 
As $k\to\infty$ the sequence of iterates \eqref{eq:batch_prox} provably converges to a minimizer $\hbS$ [cf. \eqref{eq:opt_batch}]; see e.g.,~\cite{bauschke2011convex}. Moreover, $\varphi(\bbS^k) - \varphi(\hbS) \rightarrow 0$ due to the continuity of the composite objective function $\varphi$.


\section{Joint digraph and filter identification}\label{S:Joint_Estimation}

The shift-invariance identity $\bbH\bbS=\bbS\bbH$ captures a fundamental relationship between the filter and the GSO in the diffusion-based generative mechanism \eqref{E:Filter_input_output_time}. While so far only leveraged to recover the graph topology given a filter estimate [cf. the constraint in \eqref{E:robust_general_problem}], it is prudent to close the loop and feed back GSO estimates to also aid the initial filter identification step in Section \ref{S:EstimatingDirectedFilterFromQuadratic}. Following this rationale, here we propose a \emph{joint} digraph and filter identification framework.

From the insights gained in Sections \ref{S:EstimatingDirectedFilterFromQuadratic} and \ref{S:EstimatingDirectedShiftFromFilter}, it is only natural to combine the estimators \eqref{E:general_problem_non_symmetric_shifts} and \eqref{E:robust_general_problem} to arrive at 
\begin{align}\label{E:joint_problem_non_symmetric_shifts} 
\min_{\bbS,\bbH,\{\bbU_m\}} &\
\|\bbS\|_1 +\frac{\lambda}{2}\sum_{m=1}^{M}\| \bbH-\hbC_{\bby,m}^{1/2} \bbU_m \bbC_{\bbx,m}^{-1/2} \|_F^2  \nonumber \\
\text{s. to }\;
&\bbU_m\in\ccalU_N,\,\:\:m=1,\ldots,M,\\ 
&\bbS \in \ccalS,  \quad
\| \bbH \bbS - \bbS \bbH \|_{F}^2 \leq \varepsilon.\nonumber
\end{align}
The solution to the nonconvex problem \eqref{E:joint_problem_non_symmetric_shifts} offers a one-shot digraph estimator $\hbS$, where one can treat the filter-related variables $\bbH$ and $\{\bbU_m\}_{m=1}^M$ as nuisance parameters. 

We again use IAM to develop a digraph topology inference algorithm, splitting the variables in three groups $\bbS$, $\{\bbU_m\}$, and $\bbH$ over which we will inexactly minimize \eqref{E:joint_problem_non_symmetric_shifts} in a cyclic fashion. The good news is that the resulting subproblems are closely related to \eqref{E:general_problem_non_symmetric_shifts} and \eqref{eq:opt_batch}. At iteration $k=0,1,2,\ldots$ we minimize \eqref{E:joint_problem_non_symmetric_shifts} first w.r.t. $\bbS$, while $\bbH^k$ and $\{\bbU_m^k\}$ are kept fixed to their most recent values. If we dualize the constraint $\|\bbH^k \bbS - \bbS \bbH^k\|_F^2\leq \varepsilon$ as in Section \ref{Ss:Batch_Proximal}, then we find that $\bbS^{k+1}$ is given by the solution of \eqref{eq:opt_batch} after substituting $\hbH \leftarrow \bbH^k$. Rather than fully solving this subproblem as in Algorithm \ref{A:alg1}, we run a few inner PG iterations \eqref{eq:batch_prox} replacing \eqref{eq:grad_g} with
\begin{equation} \label{eq:grad_g_modified}
	\nabla g(\bbS) =\! \mu\left[\!\left(\bbH^k\right)^{\!\top}\!(\bbH^k \bbS - \bbS \bbH^k) - (\bbH^k \bbS - \bbS \bbH^k)\left(\bbH^k\right)^{\!\!\top}\right]{\!}.
\end{equation}

Having updated $\bbS^{k+1}$, we proceed to inexactly minimize \eqref{E:joint_problem_non_symmetric_shifts} w.r.t. the remaining variables $\{\bbH,\{\bbU_m\}\}$, namely
\begin{align}\label{E:filter_subproblem} 
\min_{ \bbH,\{\bbU_m\}} \; &\frac{\mu}{2}\| \bbH \bbS^{k+1} - \bbS^{k+1} \bbH \|_{F}^2\nonumber \\
&+\frac{\lambda}{2}\sum_{m=1}^{M}\| \bbH-\hbC_{\bby,m}^{1/2} \bbU_m \bbC_{\bbx,m}^{-1/2} \|_F^2 \nonumber \\
\text{s. to }\;
&\bbU_m\in\ccalU_N,\,\:\:m=1,\ldots,M, 
\end{align}
which closely resembles \eqref{E:general_problem_non_symmetric_shifts}, but for the additional shift-invariance regularization in the objective. The importance of this regularization term cannot be overstated. First, it captures the closed-loop nature of the filter estimator (cf. Fig. \ref{F:scheme_topo_id}), which now benefits from the most recent GSO estimate. This is a direct consequence of the joint formulation \eqref{E:joint_problem_non_symmetric_shifts}; setting $\mu=0$ in \eqref{E:filter_subproblem} yields the open loop estimator in \eqref{E:general_problem_non_symmetric_shifts}. 

\begin{algorithm}[t]
	\caption{IAM for digraph topology identification}
	\label{alg:topoid_directed_inexact}
	\begin{algorithmic}[1]
		\REQUIRE $\{\bbC_{\bbx,m}\}_{m=1}^{M}$, $\{\hbC_{\bby,m}\}_{m=1}^{M}$, parameters $t_{\max},\epsilon> 0$.
		\STATE \textbf{initialize} $k=0$, $\{\bbU_m^0\}_{m=1}^{M} \in\ccalU_N$, $\bbH^0$, $\bbS^0$  at random.
        \REPEAT 
        \STATE  \textbf{initialize} $t=0$, $\bbS^0 := \bbS^k$, and $\gamma=1/[4\mu\sigma_{\max}^2(\bbH^k)]$.
		\REPEAT 
        \STATE Compute $\nabla g(\bbS^t)$ using \eqref{eq:grad_g_modified}. 
		\STATE Take GD step $\bbD^t=\bbS^t-\gamma \nabla g(\bbS^t)$. 
		\STATE Update $\bbS^{t+1} = \text{prox}_{\gamma \| \cdot \|_1,\ccalS}\left(\bbD^t \right)$ via \eqref{eq:prox_S_P}.
	    \STATE  $t \gets t+1$.
		\UNTIL{$t=t_{\max}$}.
		\STATE Update $\bbS^{k+1} := \bbS^t$.
		\FOR{$m=1,\ldots,M$ in parallel}
		\STATE  \textbf{initialize} $t=0$ and $\bbU^0 := \bbU_m^k$.
		\REPEAT
		\STATE Compute $\bbG_{m}^{t}$ using \eqref{e:nabla_fm}.
		\STATE Compute $\grad f_{\bbU_m}(\bbU^t)$ using \eqref{e:grad_fm}.
        \STATE Choose step size $\alpha$ via the Armijo rule.
        \STATE Update $\bbU^{t+1}$ via Riemannian GD step in \eqref{eq:r_grad_step}.
		\STATE  $t \gets t+1$.
		\UNTIL{$t=t_{\max}$}.
		\STATE Update $\bbU_m^{k+1} := \bbU^t$.
		\ENDFOR
		\STATE  \textbf{initialize} $t=0$ and $\bbH^0 := \bbH^k$.
		\REPEAT 
        \STATE Compute $\nabla h(\bbH^t)$ using \eqref{eq:gradient_wrt_filter_jointproblem}. 
        \STATE Choose step size $\beta$ via the Armijo rule.
		\STATE Update $\bbH^{t+1}=\bbH^t-\beta \nabla h(\bbH^t)$. 
	    \STATE  $t \gets t+1$.
		\UNTIL{$t=t_{\max}$}.
		\STATE Update $\bbH^{k+1} := \bbH^t$.
		\STATE $k = k + 1$.
        \UNTIL{$\| \bbS^{k} - \bbS^{k-1}  \|_{F} / \| \bbS^{k-1} \|_{F} \leq \epsilon$}.
		\RETURN $\hbS = \bbS^k$.
	\end{algorithmic}
\end{algorithm}

We update $\bbH^{k+1}$ and $\{\bbU_m^{k+1}\}$ in \eqref{E:filter_subproblem} with a minor modification to Algorithm \ref{alg:topoid_directed}. Instead of using \eqref{E:filter_estimate} to update the filter in closed form, we run a few inner GD iterations with step size $\beta>0$. Writing the objective of \eqref{E:filter_subproblem} as $h(\bbH)$ when viewing $\{\bbS^{k+1},\{\bbU_m^{k+1}\}\}$ as fixed parameters, we have
\begin{align}\label{eq:gradient_wrt_filter_jointproblem}
    \nabla_\bbH h(\bbH\})=&\mu (\bbH\bbS\bbS^\top + \bbS^\top\bbS\bbH 
    - \bbS\bbH\bbS^\top - \bbS^\top\bbH\bbS) \nonumber \\ & + \lambda \Big(M\bbH - \sum_{m=1}^M \hbC_{\bby,m}^{1/2} \bbU_m \bbC_{\bbx,m}^{-1/2}\Big) 
\end{align}
where with a slight abuse of notation, we have dropped the superindex ${k+1}$ in $\bbS$ and the $\{\bbU_m\}$. Granted, one can solve $\nabla_\bbH h(\bbH\})=\mathbf{0}$
in closed form using matrix vectorization techniques and inverting an $N^2\times N^2$ matrix, but the complexity of said inversion justifies our inexact GD update. Once more, step size $\beta>0$ can be chosen adaptively using the Armijo rule or based on the reciprocal of the Lipschitz constant $L_{\bbH}=4\mu\sigma^2_{\max}(\bbS)+\lambda M$ of $\nabla_\bbH h$. The overall IAM method is tabulated under Algorithm \ref{alg:topoid_directed_inexact} and its complexity is also $\ccalO(N^3)$.

All in all, Algorithm \ref{alg:topoid_directed_inexact} computes  inexact solutions of the mutually-reinforcing subproblems in an alternating fashion. This repeated improvement procedure
converges to a stationary solution of the joint digraph and filter identification problem \eqref{E:joint_problem_non_symmetric_shifts}; the subject dealt with next.

\subsection{Convergence Analysis}\label{Ss:Convergence}

Algorithm~\ref{alg:topoid_directed_inexact} is an IAM instance, where each block is refined by a finite number of inner iterations: a PG sweep on $\bbS$ [cf.~\eqref{eq:batch_prox}], Riemannian gradient steps with polar retraction on each $\bbU_m$ [cf.~\eqref{eq:r_grad_step}], and Euclidean gradient steps on $\bbH$ [cf.~\eqref{eq:gradient_wrt_filter_jointproblem}]. Establishing convergence is non-trivial because the objective couples a \emph{nonsmooth} term $\|\bbS\|_1$, a \emph{nonconvex} commutator coupling, and \emph{orthogonality} constraints. Off-the-shelf proximal-alternating (PALM)~\cite{BST2014} and block-coordinate descent (BCD)~\cite{peng2023bcdmanifold} guarantees do not apply to this setting.

\begin{figure*}[!t]
    \centering
    \begin{subfigure}[b]{0.45\textwidth}
        \includegraphics[width=\textwidth]{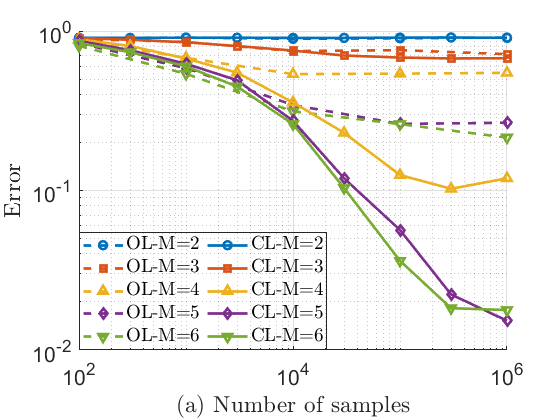}
    \end{subfigure}
    \begin{subfigure}[b]{0.45\textwidth}
        \includegraphics[width=\textwidth]{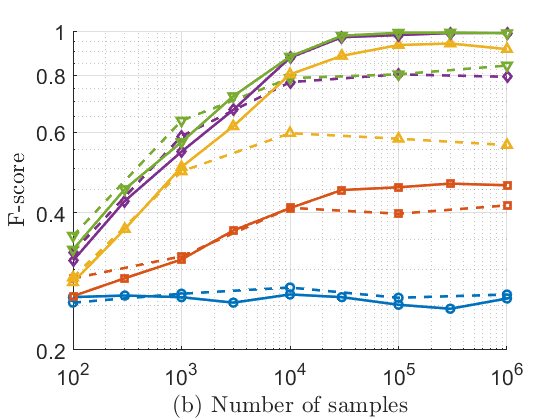}
    \end{subfigure}
    \caption{Performance of the proposed open loop (OL) and closed loop (CL) approaches, in terms of (a) estimation error and (b) F-score while varying the number of available samples $P$ for different numbers of input–output processes $M \in \{2,3,4,5,6\}$.}
    \label{fig:exp1_SJ}
    \vspace{-0.5cm}
\end{figure*}

We analyze the penalized counterpart of \eqref{E:joint_problem_non_symmetric_shifts} that Algorithm~\ref{alg:topoid_directed_inexact} actually solves, namely 
\begin{equation}\label{E:joint_penalized}
	\min_{\bbS\in\ccalS,\,\bbH,\,\{\bbU_m\}\in\ccalU_N^M} \hspace{-0.8cm}
	F(\bbS,\bbH,\{\bbU_m\}):=\|\bbS\|_1 + h(\bbS,\bbH,\{\bbU_m\}),
\end{equation}
where the smooth component $h(\bbS,\bbH,\{\bbU_m\})$ is the objective in \eqref{E:filter_subproblem} with $\bbS$ also viewed as an optimization variable. 
%
%
We invoke the following mild assumptions:
\begin{itemize}
	\item[\textbf{(A1)}] Set $\ccalS$ is nonempty, closed, convex, and semialgebraic. 
	\item[\textbf{(A2)}] Parameters $\mu,\lambda>0$ and matrices $\hbC_{\bby,m}^{1/2}$, $\bbC_{\bbx,m}^{-1/2}$ are fixed and of finite norm (the empirical covariances are full rank). Consequently $h$ is a fixed multivariate polynomial, hence $C^\infty$ with gradient Lipschitz on bounded sets.
	\item[\textbf{(A3)}] The number of inner iterations per block ($t_{\max}\geq 1$) is finite, 
    and each block update uses either a constant step size below the corresponding (inverse) Lipschitz threshold or a backtracking line search (e.g., the Armijo rule) enforcing the block sufficient-decrease inequality. 
\end{itemize}
Collecting the variables in $\bbZ:=\{\bbS,\bbH,\{\bbU_m\}_{m=1}^M\}$, we call $\bbZ^\star$ a \emph{critical point} of $F$ if $\bbS^\star$ is prox-stationary, i.e., $-\nabla_{\bbS} h(\bbZ^\star)\in\partial\big(\|\cdot\|_1+\iota_\ccalS\big)(\bbS^\star)$; $\nabla_{\bbH} h(\bbZ^\star)=\bbzero$; and each $\bbU_m^\star$ is Riemannian-stationary on $\ccalU_N$, i.e., $\grad_{\bbU_m} h(\bbZ^\star)=\bbzero$ [cf.~\eqref{e:grad_fm} where $h$ simplifies to $\phi$]. We have the following result.

\begin{mytheorem}\label{T:alg_convergence}
	Let $\{\bbZ^k\}_{k\geq 0}$ be the outer iterates generated by Algorithm~\ref{alg:topoid_directed_inexact} under Assumptions \textnormal{A1-A3}. Then:
	\begin{enumerate}
		\item[\textnormal{(i)}] Values $\{F(\bbZ^k)\}$ are non-increasing and convergent, and the iterates remain in a compact set;
		\item[\textnormal{(ii)}] Successive increments are square-summable, so $\|\bbZ^{k+1}-\bbZ^k\|_F\to 0$ and every accumulation point of $\{\bbZ^k\}$ is a critical point of $F$; and
		\item[\textnormal{(iii)}] Sequence $\{\bbZ^k\}$ has finite length, $\sum_{k}\|\bbZ^{k+1}-\bbZ^k\|_F<\infty$, and converges to a single critical point $\bbZ^\star$ of $F$.
	\end{enumerate}
\end{mytheorem}

The assumptions are mild and hold throughout our experiments in Section \ref{S:Simulations}: the admissible adjacency set $\ccalS$ in Section~\ref{S:EstimatingDirectedShiftFromFilter} is a polyhedron (hence closed, convex, and semialgebraic), full-rank empirical covariances make $\hbC_{\bby,m}^{1/2}$ and $\bbC_{\bbx,m}^{-1/2}$ well defined, and admissible step sizes follow either from Lipschitz constants such as $L_{\bbS}=4\mu\sigma_{\max}^2(\bbH^k)$ 
(cf.~Section~\ref{Ss:Batch_Proximal}) or backtracking. Three points are noteworthy. First, the guarantees are \emph{independent} of the inner-iteration counts and of the cyclic order in which the blocks are updated; these affect only the constants in the proof, so the $\bbS\!\to\!\{\bbU_m\}\!\to\!\bbH$ schedule of Algorithm~\ref{alg:topoid_directed_inexact} can be modified if deemed convenient. Second, despite the nonconvexity of \eqref{E:joint_penalized}, the limit is guaranteed to be stationary, not merely a fixed point of the iteration map. Third, to the best of our knowledge this is the first convergence result for a digraph topology-inference objective that simultaneously involves an $\ell_1$-norm for sparsity as well as a manifold-constrained filter-identification term; existing analyses~\cite{peng2023bcdmanifold} do not apply precisely because of the nonsmooth $\|\bbS\|_1$ on the proximal block. The proof relies on a sufficient-decrease inequality paired with a relative-error (subgradient) bound, enabling a Kurdyka--\L ojasiewicz (KL) argument. Since $F$ is semialgebraic it satisfies the KL condition everywhere, and this is what upgrades subsequential convergence to convergence of the entire $\{\bbZ^k\}_{k\geq 0}$. A complete proof, explicit step-size conditions, and KL-exponent-dependent rates are provided in the supplementary material.

\section{Numerical Performance Evaluation}\label{S:Simulations}

We conduct numerical experiments to evaluate the performance of the digraph topology inference algorithms under different synthetic scenarios and also using real-world data.

\subsection{Simulated Experiments}
\label{ssec:synthetic_data}

\noindent\textbf{Performance under different $M$ values.}
This test case probes the ability of the proposed methods to recover a digraph from $M$ input-output network processes under limited sample availability for covariance estimation. We consider Erd\H{o}s--R\'enyi (ER) random digraphs with $N=20$ nodes, where each directed edge is drawn independently with probability $p=0.3$. The graph filter is generated as
$
\bbH_1 = \sum_{l=1}^{L} h_l \bbS^{\,l-1},$ i.i.d.
$h_l \sim \textrm{Normal}(0,1)$
and order $L=3$. For each graph realization, we simulate $M$ input--output covariance pairs $\{\bbC_{\bbx,m},\hbC_{\bby,m}\}_{m=1}^M$. We let $\bbC_{\bbx,m} = \bbB_m \bbB_m^\top$, where the entries of $\bbB_m$ are i.i.d. from a standard normal distribution. We sample $P$ i.i.d. inputs $\bbx_m^{(p)}\sim\textrm{Normal}(\mathbf{0},\bbC_{\bbx,m})$ for each $m$, and let $\bby_m^{(p)}=\bbH_1\bbx_m^{(p)}$. The corresponding $\hbC_{\bby,m}$ are obtained through sample averaging as in \eqref{E:cov_estimate}. 
The number of available covariance pairs varies from $M=2$ to $M=6$, and all results are averaged over $100$ independent graph realizations.

\begin{figure*}[t]
    \centering
    \begin{subfigure}[b]{0.45\textwidth}
        \includegraphics[width=\textwidth]{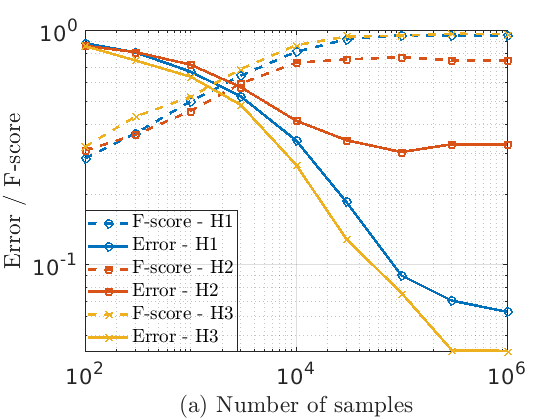}
    \end{subfigure}
    \begin{subfigure}[b]{0.45\textwidth}
        \includegraphics[width=\textwidth]{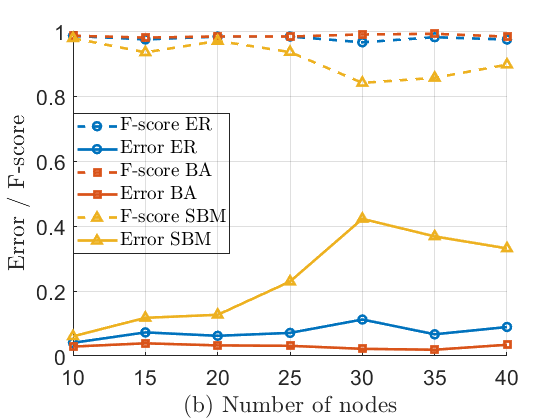}
    \end{subfigure}
    \caption{Estimation error and F-score under different experimental conditions: (a) varying the number of samples $P$ for different graph filters $\{\bbH_1, \bbH_2, \bbH_3\}$; and (b) varying the number of nodes $N$ for different random graph models $\{\text{ER}, \text{BA}, \text{SBM}\}$.}
    \label{fig:sinth_exp}
    \vspace{-0.5cm}
\end{figure*}

We compare the proposed estimation strategies in Fig. \ref{F:scheme_topo_id}. The two-step open-loop (OL) estimator first identifies the graph filter $\bbH$ and the GSO $\bbS$ sequentially by running Algorithms \ref{alg:topoid_directed} and \ref{A:alg1} without feeding the graph estimate back into the filter-identification step. The closed-loop (CL) estimator corresponds to the joint formulation introduced in Section~\ref{S:Joint_Estimation}, where $\bbH$ and $\bbS$ are estimated simultaneously using Algorithm~\ref{alg:topoid_directed_inexact}. 

Fig.~\ref{fig:exp1_SJ}(a) reports the graph estimation $\textrm{Error}=\|\hbS-\bbS\|_F^2/\|\bbS\|_F^2$ as a function of the number of samples $P$ per process. For small numbers of covariance pairs, namely $M \in \{2,3\}$, both schemes exhibit poor recovery performance even in the large-sample regime, illustrating the intrinsic difficulty of the problem and highlighting the importance of having multiple input--output covariance observations for reliable graph identification. As $M$ increases, both methods improve, but the gain is markedly larger for the the joint formulation. In particular, the CL method with $M=4$ already achieves estimation errors comparable to those obtained by the OL approach with $M \in \{5,6\}$, and even outperforms it in the large-sample regime. Moreover, going from $M=4$ to $M=5$ shows the CL scheme can match its OL counterpart's estimation accuracy with two orders of magnitude fewer samples.

Fig.~\ref{fig:exp1_SJ}(b) presents the corresponding F-score values for graph support recovery, namely the harmonic mean of edge recovery precision and recall. The CL approach generally achieves higher F-scores than the OL method across most values of $M$, with comparable performance only for $M=2$. Furthermore, for sufficiently large $P$ and $M \in \{5,6\}$, the CL method achieves nearly perfect support recovery, whereas the OL approach saturates around an F-score of $0.8$. Overall, these results provide two main insights. First, they are consistent with the identifiability claims of Theorem~\ref{T:new_uniqueness}, confirming that input diversity plays a critical role towards accurate digraph estimation. Second, they demonstrate the benefit of jointly estimating $\bbS$ and $\bbH$, as the structural coupling imposed by the joint formulation (cf. the relaxed conmutation constraint) leads to significantly improved graph recovery performance.\vspace{2pt}

\noindent\textbf{Performance under different graph filters.}
We test the robustness of Algorithm~\ref{alg:topoid_directed_inexact} under different diffusion filter structures. For $L=3$, we consider three graph filter models: 
\begin{itemize}
    \item $\bbH_1 = \sum_{l=1}^{L} h_l \bbS^{\,l-1}$, with i.i.d. $h_l \sim \textrm{Normal}(0,1)$;
    \item $\bbH_2 = (\bbI + \alpha \bbS)^{-1}$, with $\alpha \sim \textrm{Uniform}(0,1)$ and;
    \item $\bbH_3 = \sum_{l=1}^{L} h_l \bbS^{\,l-1}$, with i.i.d. $h_l \sim \mathrm{Exp}(2)$, sorted in descending order and normalized such that $\max_l h_l = 1$.
\end{itemize}
The three models induce fundamentally different spectral and structural behaviors. The Gaussian-tap filter $\bbH_1$ produces a generic linear combination of GSO powers, which may include positive and negative contributions that partially cancel across orders, increasing variability across realizations. The exponential filter $\bbH_3$, in contrast, enforces strictly positive and monotonically decaying coefficients. Accordingly, most of the energy is concentrated on low-order shifts and it yields a more stable and better-conditioned mapping from inputs to outputs, improving identifiability of the underlying digraph. Finally, $\bbH_2$ induces a global diffusion mechanism where all nodes interact through infinitely many powers of $\bbS$. This rational form typically leads to stronger coupling and higher sensitivity to noise, making the inverse problem more ill-conditioned than in the finite-order cases.

\begin{figure*}[t]
    \centering

    \begin{subfigure}[b]{0.32\textwidth}
        \includegraphics[width=\textwidth]{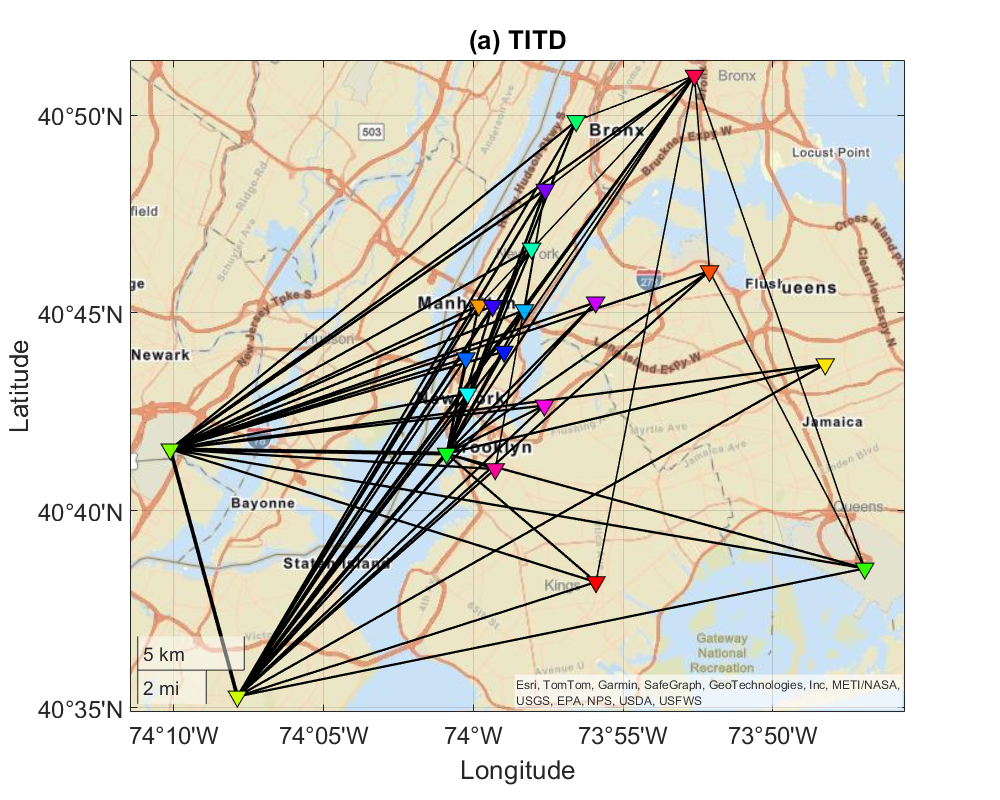}
    \end{subfigure}
    \begin{subfigure}[b]{0.32\textwidth}
        \includegraphics[width=\textwidth]{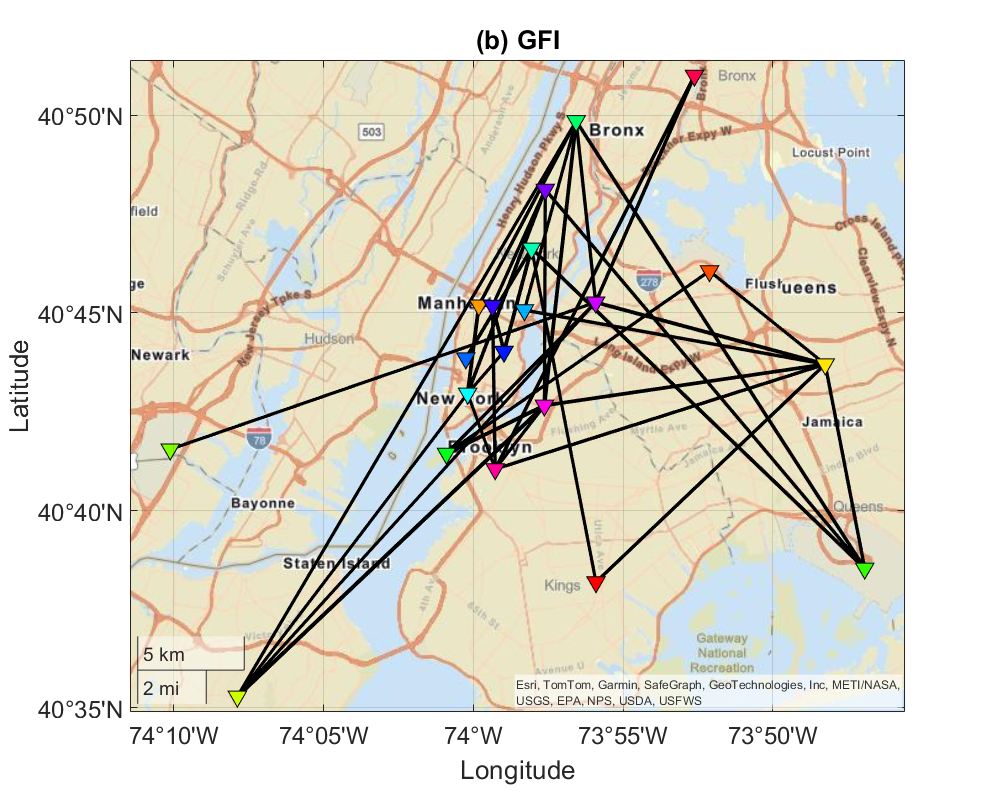}
    \end{subfigure}
    \begin{subfigure}[b]{0.32\textwidth}
        \includegraphics[width=\textwidth]{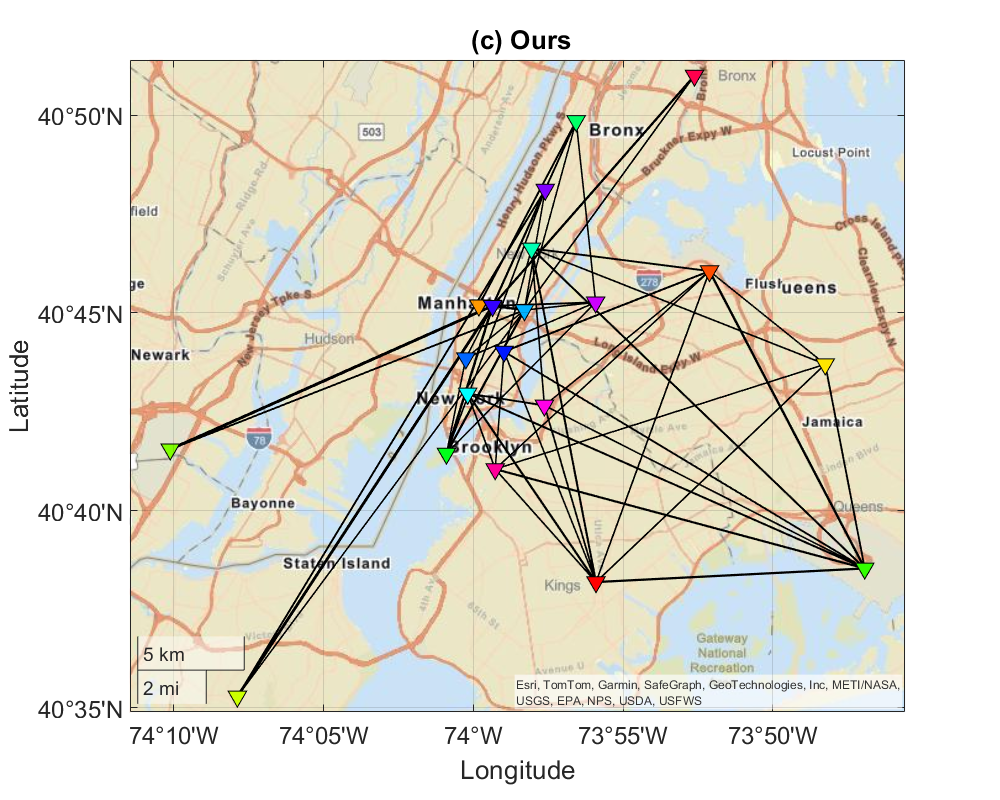}
    \end{subfigure}

    \vspace{0.1cm}

    \begin{subfigure}[b]{0.32\textwidth}
        \includegraphics[width=\textwidth]{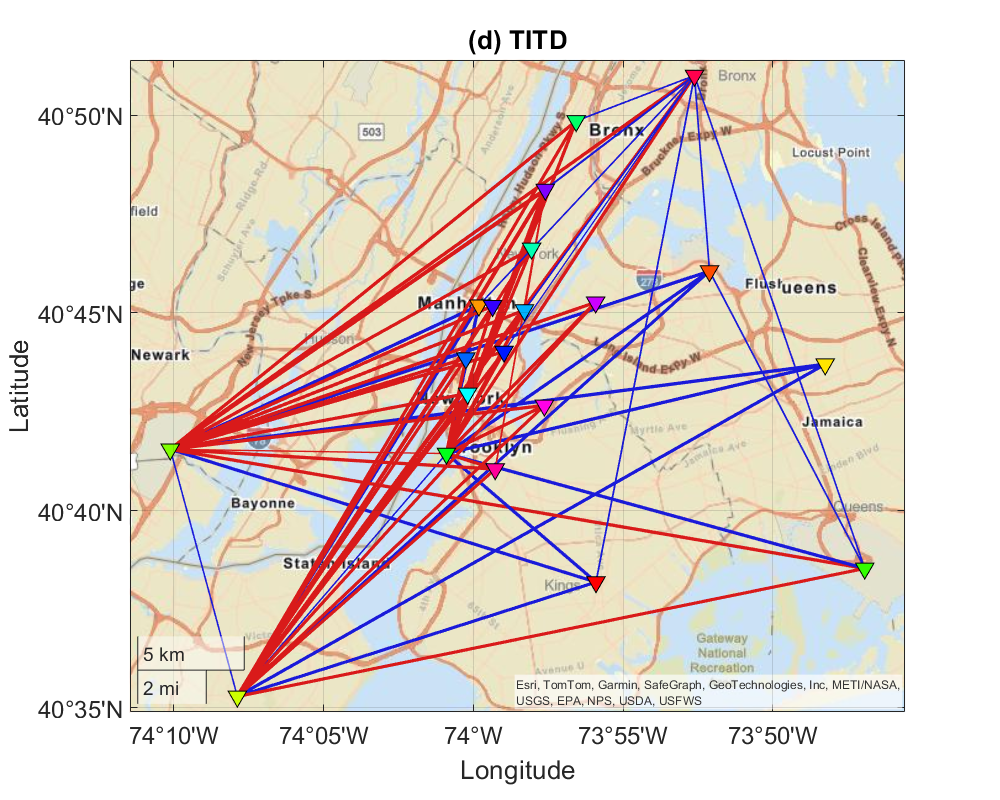}
    \end{subfigure}
    \begin{subfigure}[b]{0.32\textwidth}
        \includegraphics[width=\textwidth]{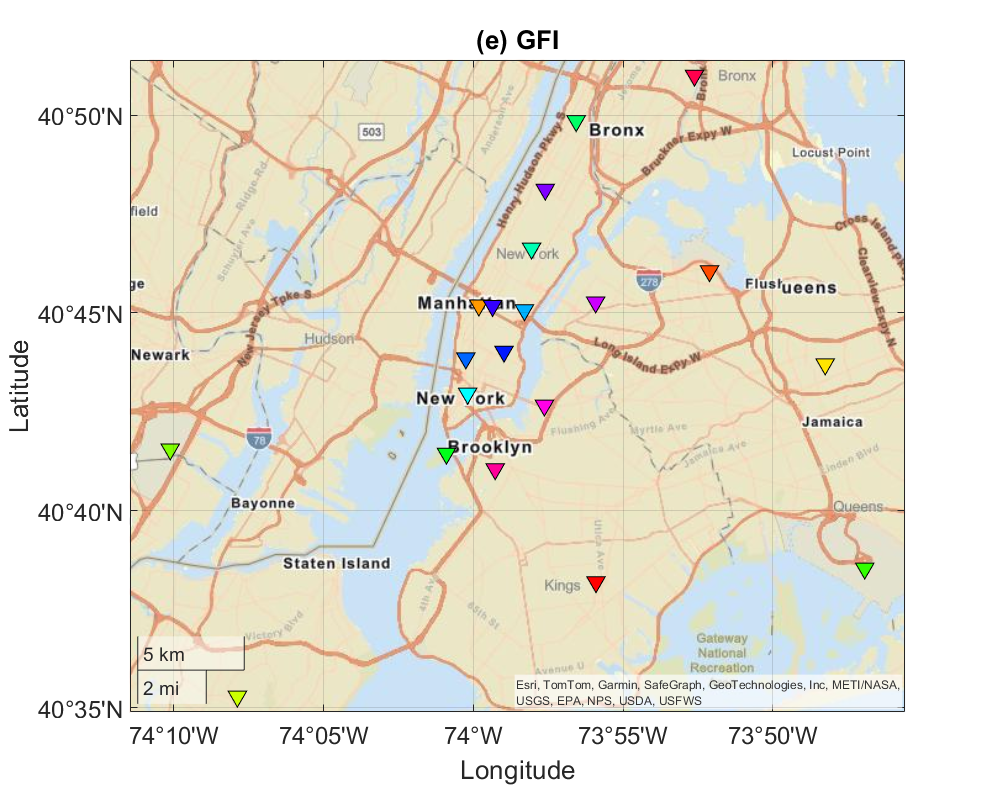}
    \end{subfigure}
    \begin{subfigure}[b]{0.32\textwidth}
        \includegraphics[width=\textwidth]{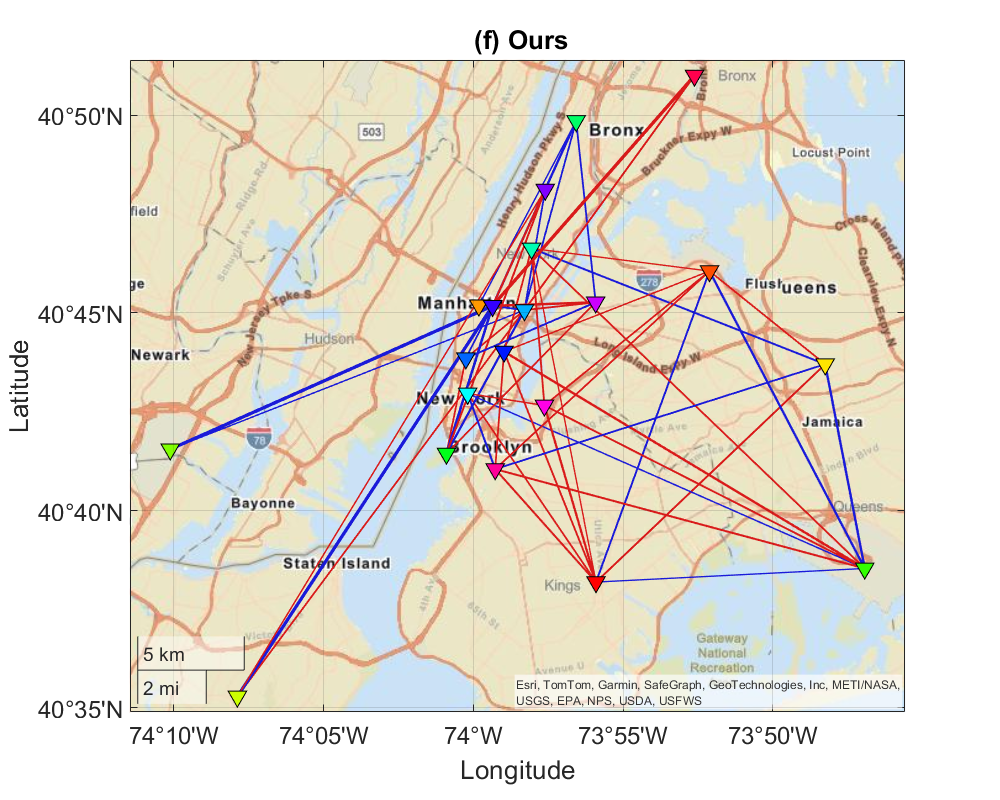}
    \end{subfigure}
    \caption{The columns show the estimated mobility links of three different approaches: TITD~\cite{shen2017tensors}, GFI~\cite{RSSSAMGM_OJSP21}, and Algorithm ~\ref{alg:topoid_directed_inexact}. The second row highlights directional differences for the digraphs, where line width represents traffic intensity.}
    \label{fig:uber_exp}
    \vspace{-0.5cm}
\end{figure*}

Fig.~\ref{fig:sinth_exp}(a) reports the estimation Error and F-score as functions of the number of samples $P$ per process, with $M=4$ fixed. As expected, increasing $P$ improves both metrics across all filter types. In terms of relative performance, $\bbH_3$ consistently achieves the lowest estimation error and highest F-score, followed by $\bbH_1$, while $\bbH_2$ yields the most challenging recovery. This ordering is consistent with the conditioning induced by each filter. The structured decay of $\bbH_3$ improves numerical stability and reduces ambiguity in separating contributions of different GSO orders, whereas the inverse form of $\bbH_2$ amplifies global dependencies and increases the sensitivity of the covariance mapping to estimation noise. These results confirm that the proposed method remains effective across heterogeneous filter classes, with performance degradation directly linked to the intrinsic conditioning and locality properties of the graph filter.\vspace{2pt} 

\noindent\textbf{Performance across random graph models.} Here we examine recovery performance under several random digraph ensembles. Specifically, we consider three classes of directed networks: ER, Barab\'asi--Albert (BA), and stochastic block model (SBM) graphs with $3$ communities. Consistent with the previous experimental setup, we simulate $M=4$ independent input--output processes and utilize $P=10^6$ samples per process. The generative diffusion filter is fixed to $\bbH_3$.

Fig.~\ref{fig:sinth_exp}(b) depicts the graph estimation Error and F-score as a function of the number of nodes $N$. When comparing the recovery performance across the different topologies (parameters are chosen to yield comparable edge densities in each case), the results demonstrate a clear performance advantage for the BA and ER models over the SBM. For ER and BA graphs, Algorithm~\ref{alg:topoid_directed_inexact} maintains an F-score near $1$ and a consistently low estimation error across all evaluated network sizes, with the BA graphs yielding marginally superior accuracy.

Conversely, topology inference over SBM graphs experiences a noticeable degradation in performance as $N$ grows. While the recovery metrics remain competitive for smaller graphs, Error markedly increases and the F-score deteriorates as the number of nodes grows from $N=20$ to $40$. These findings suggest that while the proposed method is resilient to scaling in homogeneous ER or BA network topologies, distinct structural properties such as the dense intra-community connectivity inherent to SBMs render the digraph learning task progressively more challenging as $N$ grows.\vspace{2pt}

\subsection{Real Data Experiments}
\label{ssec:real_data}

\noindent\textbf{Uber commute patterns in New York City.}
We evaluate the proposed dgraph topology inference framework using real mobility data from Uber pickups in New York City. The dataset contains the time and location of all Uber pickups between January~1 and June~29, 2015, distributed across 263 location IDs. Following the preprocessing strategy in~\cite{RSSSAMGM_OJSP21}, the locations are grouped into $N=20$ spatially contiguous regions based on geographical proximity. The centroid of each region defines a node in the latent urban mobility graph to be inferred. For every region, the total number of pickups within selected time intervals is aggregated to generate graph signals.

We consider $M=3$ mobility processes corresponding to Monday-Tuesday traffic ($m=1$), Wednesday-Thursday traffic ($m=2$), and Friday-Saturday-Sunday traffic ($m=3$). For each day, pickups collected between 5~a.m.--1~p.m.\ are treated as input signals $\mathbf{x}$, while pickups between 2--9~p.m.\ define output signals $\mathbf{y}$. Each day therefore provides one input--output observation pair $(\mathbf{x},\mathbf{y})$, assigned to the corresponding process. Using these observations, we estimate the covariance matrices $\{\hbC_{\bbx,m},\hbC_{\bby,m}\}_{m=1}^{3}$ and run Algorithm~\ref{alg:topoid_directed_inexact} to jointly recover the graph filter $\widehat{\bbH}$ and the underlying sparse directed mobility graph. This formulation models urban population movement as a diffusion process evolving over a latent transportation network, where morning commute propagates toward afternoon mobility patterns. Although simplified, this linear diffusion model offers an approximation for large-scale aggregate mobility dynamics and has been adopted in graph-based transportation analysis~\cite{thanou17,RSSSAMGM_OJSP21}.

Figs.~\ref{fig:uber_exp}(a)--(c) compare the mobility graphs inferred by three different approaches: (i) the digraph topology identification via tensor decomposition (TITD) method from~\cite{shen2017tensors}; (ii) the \emph{undirected} graph filter identification (GFI) method from~\cite{RSSSAMGM_OJSP21}; and (iii) the proposed Algorithm~\ref{alg:topoid_directed_inexact}. The graph estimated by TITD in Fig.~\ref{fig:uber_exp}(a) exhibits a highly centralized structure dominated by three regions, namely Newark Airport, Blood Root Valley, and Governors Island, which concentrate most of the connectivity while the remaining areas are only weakly interconnected. Such star-like connectivity patterns are difficult to reconcile with realistic urban transportation dynamics, where mobility is typically distributed across multiple densely connected regions. The GFI result in Fig.~\ref{fig:uber_exp}(b) produces a substantially more structured topology, where several Manhattan districts become strongly interconnected and a more balanced connectivity pattern is observed across the remaining regions. The graph inferred by the proposed method in Fig.~\ref{fig:uber_exp}(c) preserves these large-scale connectivity patterns while additionally introducing edge directionality, thus capturing asymmetric mobility interactions between urban areas.

\begin{table}
\centering
\caption{Total return (\%) for different portfolios under inverse degree and inverse eigenvector centrality weighting.}
\begin{tabular}{lcc}
\toprule
\textbf{Portfolio} & \textbf{Inverse Degree} & \textbf{Inverse Eigenvector Centrality} \\
\midrule
GFI & 24.00  & 24.33 \\
TITD    & 25.46  & 16.53 \\
Ours      & \textbf{39.13}  & \textbf{35.01} \\
Uniform   & 25.20  & 25.20 \\
\bottomrule
\end{tabular}\vspace{-0.3cm}
\end{table}

To better visualize directional mobility behavior, Figs.~\ref{fig:uber_exp}(d)--(f) depict the directional imbalance between each pair of connected nodes. Red edges indicate dominant flows from node $i$ to node $j$ with $i<j$, whereas blue edges indicate dominant flows in the opposite direction. Edge thickness encodes the magnitude of the directional asymmetry. Since GFI recovers an undirected graph, it cannot represent directional traffic patterns; consequently, Fig.~\ref{fig:uber_exp}(e) shows each one of the 20 centroids associated with each region and no links between them because the graph is symmetric. In contrast, the proposed method reveals interpretable asymmetric mobility flows. Based on the predefined node ordering, the graph captures distinct outbound traffic patterns from Manhattan. In particular, we observe dominant flows moving away from Manhattan towards outer residential boroughs and the three airports. For instance, the blue edges connecting Manhattan regions ($j>6,7$) to Newark Airport (node 6) and JFK Airport (node 7) indicate concentrated traffic from Manhattan towards these airports. Similarly, the Bronx (node 20) and Staten Island (node 5) receive incoming traffic from lower-indexed central nodes, represented by red edges ($i \to 20$) and blue edges ($j \to 5$) respectively. This behavior is consistent with typical afternoon or evening commute patterns in New York City, where outbound trips from business districts to residential areas and evening airport departures peak. Interestingly, LaGuardia Airport (node 2) displays the opposite trend, with red edges indicating outgoing flow towards Manhattan ($2 \to j$). These results show that the proposed method not only recovers a richer mobility topology than TITD by considering general filters others than the one induced by a linear SEM, but also captures complex, asymmetric traffic flows between regions, which cannot be identified by undirected graph models.\vspace{2pt}

\noindent\textbf{Portfolio optimization.} Finally, we test the proposed framework in a financial portfolio optimization task using stock price data from seven major companies. The experiment consists of two stages: (i) digraph estimation from historical market data; and (ii) portfolio construction based on the inferred financial dependency graphs. During the training stage, daily \textit{open} and \textit{close} prices were collected from June~3,~2019, to April~30,~2020, covering 231 trading days. From these observations, covariance matrices of open prices ($\bbC_\bbx$) and close prices ($\bbC_\bby$) were estimated at three temporal resolutions: daily, weekly, and monthly. These covariance matrices were then used to infer financial interaction graphs using three different approaches: (i) GFI~\cite{RSSSAMGM_OJSP21}, (ii) TITD~\cite{shen2017tensors}, and (iii) Algorithm~\ref{alg:topoid_directed_inexact}. Since no ground-truth financial topology is available, the inferred graphs are interpreted as latent dependency structures encoding relations between assets.

\begin{figure}[t]
    \centering
    \begin{subfigure}[b]{1\columnwidth}
        \centering
        \includegraphics[width=\textwidth]{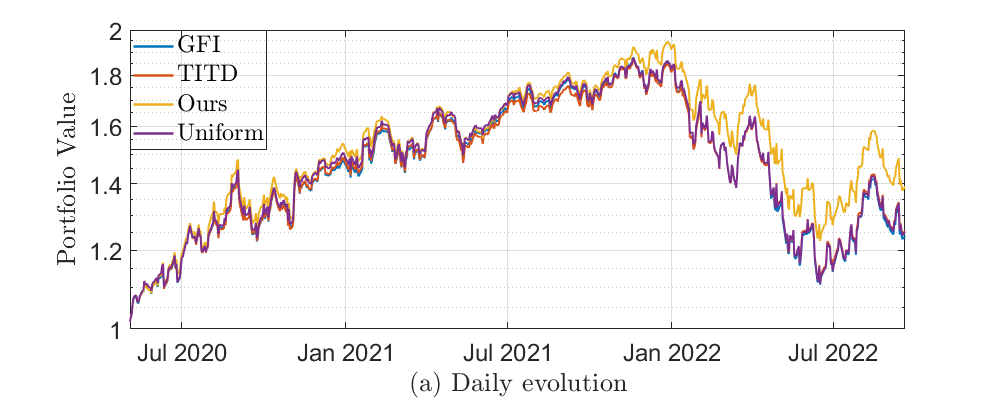}
    \end{subfigure}

    \vspace{0.25cm}

    \begin{subfigure}[b]{1\columnwidth}
        \centering
        \includegraphics[width=\textwidth]{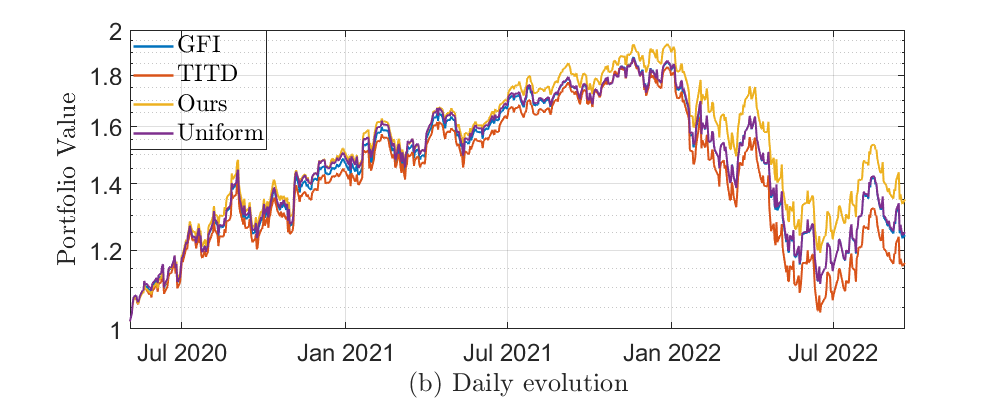}
    \end{subfigure}

    \caption{Portfolio performance obtained by considering four different weights for the portfolio optimization from the estimated graphs by considering: (i) inverse degree; and (ii) inverse eigenvector centrality allocation.}
    \label{fig:financial_exp}
    \vspace{-0.3cm}
\end{figure}

The quality of the estimated digraphs is assessed indirectly through their usefulness in portfolio construction. In the testing stage, we used market data from May~4,~2020, to September~19,~2022, and computed daily log-returns from the corresponding \textit{open} and \textit{close} prices. Portfolio weights were derived from the inferred graph topologies using two graph-based allocation strategies: (i) inverse degree weighting, which assigns larger weights to less connected assets in order to promote diversification; and (ii) inverse eigenvector centrality weighting, which reduces the exposure to highly influential assets to mitigate concentration and systemic risk. In addition, we consider a graph-agnostic uniform allocation baseline that assigns identical weights to all assets. For each strategy, a unit initial investment is distributed according to a normalized weight vector $\bbw$, and portfolio performance is evaluated under a buy-and-hold policy through the final accumulated portfolio value over the testing period.

Figs.~\ref{fig:financial_exp}(a) and (b) depict the portfolio value evolution under inverse-degree and inverse-eigenvector centrality allocations, respectively, while Table~I summarizes the final returns. Portfolios relying on the proposed digraph inference method consistently achieve the highest returns across all strategies and temporal resolutions. 
Conversely, the GFI and TITD baselines perform similarly to a naive uniform allocation under the inverse-degree strategy, and worse under the inverse-eigenvector centrality strategy. These results demonstrate that our framework more accurately models the underlying diffusion processes to capture latent financial dependencies. At least in this toy setting, the inferred topologies enable better portfolio diversification and superior long-term returns.


\section{Conclusions, Limitations, and Future Work}\label{S:Conclusions}

We studied the problem of inferring the topology of a digraph from observations of signals diffused on the network. Modeling the mapping between the (arbitrarily correlated) excitation inputs and the observed outputs as a generic polynomial graph filter, enabled a two-step topology inference approach whereby: i) we first use (statistical) information on the inputs and outputs to infer the said diffusion filter; and ii) we then combine the filter estimate along with prior structural information on the digraph to recover the network topology. For the system identification step, we assume only second-order statistical information on the inputs is available. While such a problem can be naturally formulated as a nonconvex fourth order optimization, we recast it as a smooth quadratic minimization subject to Stiefel manifold constraints. We established the filter is uniquely identifiable under sufficient spectral richness of the inputs and developed a convergent Riemannian gradient descent algorithm. Regarding the second step, the focus was on finding the sparsest GSO compatible with the estimated filter, namely that both operators (approximately) commute. By feeding back intermediate GSO estimates to boost filter recovery performance, we put forth a joint graph filter and topology identification formulation and associated provably convergent algorithm, closing the loop between steps i) and ii) to offer improved sample complexity. The closed-loop digraph topology inference pipeline was comprehensively evaluated using synthetic and real network data.

The problem is admittedly challenging and our solution requires fairly large sample sizes, multiple input process, and cubic (in the number of nodes) complexity solvers. To bridge these gaps, as future work we plan to investigate lightweight online algorithms, restrictions to specialized graph filter classes, and additional domain-specific constraints.

\appendix

\subsection{Proof of Lemma \ref{L:identifiability_directed}}\label{L:identifiability_directed_proof}

It follows from~\eqref{E:feas_H_non_sym_v3} that solving the system of quadratic equations in~\eqref{E:quadratic_system} amounts to finding $\{\bbU_{m}\}_{m=1}^{M}$ satisfying
\begin{equation}\label{e:H:asym:1}
\bbH \!=\! \bbC_{\bby,m}^{1/2} \bbU_m \bbC_{\bbx,m}^{-1/2},\:\:
		\bbU_m^{\top}\bbU_m = \bbI_N,\quad \text{for all} \,\, m.
\end{equation}
Equating $\bbH$ in~\eqref{e:H:asym:1} for $m$ and $m-1$ yields $\bbC_{\bby,m}^{1/2} \bbU_m \bbC_{\bbx,m}^{-1/2} = 
\bbC_{\bby,{m-1}}^{1/2} \bbU_{m-1} \bbC_{\bbx,{m-1}}^{-1/2}$. 
Since $\{\bbC_{\bbx,m}\}_{m=1}^{M}$ and $\{\bbC_{\bby,m}\}_{m=1}^{M}$ are assumed to be invertible, then
\begin{equation} \label{e:U_2}
	\bbU_{m-1} = \bbC_{\bby,{m-1}}^{-1/2}  \bbC_{\bby,m}^{1/2} \bbU_m \bbC_{\bbx,m}^{-1/2} \bbC_{\bbx,{m-1}}^{1/2}.
\end{equation}
Substituting~\eqref{e:U_2} in the orthogonality condition of~\eqref{e:H:asym:1}
\begin{align} \label{e:U_2_ortho}
	&\bbU_{m-1}^{\top}\bbU_{\m-1}  = \bbI_N = \\
	& \bbC_{\bbx,{m-1}}^{1/2} \bbC_{\bbx,m}^{-1/2} \bbU_m^\top \bbC_{\bby,m}^{1/2} \bbC_{\bby,{m-1}}^{-1}  \bbC_{\bby,m}^{1/2} \bbU_m \bbC_{\bbx,m}^{-1/2} \bbC_{\bbx,{m-1}}^{1/2}. \nonumber
\end{align}
Left and right multiplying both sides of~\eqref{e:U_2_ortho} with $\bbC_{\bbx,m}^{1/2} \bbC_{\bbx,{m-1}}^{-1/2}$ and $\bbC_{\bbx,{m-1}}^{-1/2} \bbC_{\bbx,m}^{1/2}$, respectively, and recalling the definitions in~\eqref{e:C_yyy} give rise to
\begin{equation} \label{e:U_2_ortho_simp}
	\bbU_m^\top \bbC_{\bby\bby,m} \bbU_m = \bbC_{\bbx\bbx,m}.
\end{equation}
If we show that the set of solutions of~\eqref{e:U_2_ortho_simp} is given by 
\begin{equation} \label{e:U_2_ortho_final}
	\bbU_m = \bbV_{\bby\bby,m} \text{diag}(\bbb_m) \bbV_{\bbx\bbx,m}^{\top},
\end{equation}
where $\bbb_m \in \{-1,1\}^{N}$, plugging \eqref{e:U_2_ortho_final} in~\eqref{e:H:asym:1} wraps our proof. 

In showing~\eqref{e:U_2_ortho_final}, first notice that $\bbC_{\bby\bby,m}$ and $\bbC_{\bbx\bbx,m}$ must have the same eigenvalues since they are similar matrices [cf.~\eqref{e:U_2_ortho_simp}].
Hence, a direct substitution of~\eqref{e:U_2_ortho_final} into~\eqref{e:U_2_ortho_simp} corroborates that every $\bbU_m$ as in~\eqref{e:U_2_ortho_final} solves~\eqref{e:U_2_ortho_simp}. 
Thus, we are left to show that every solution of~\eqref{e:U_2_ortho_simp} must be of the form in~\eqref{e:U_2_ortho_final}.
To see that this is the case, notice that due to matrix similarity, if $\bbv_{\bbx \bbx}$ is an eigenvector of  $\bbC_{\bbx\bbx,m}$ then $\bbU_m \bbv_{\bbx \bbx}$ must be an eigenvector of $\bbC_{\bby\bby,m}$.
We thus have $\bbV_{\bby\bby,m} = \bbU_m \bbV_{\bbx\bbx,m}$, for any choice of the eigenbasis $\bbV_{\bbx\bbx,m}$. 
Since all the eigenvalues in $\bbC_{\bbx\bbx,m}$ are distinct, the only free parameter is the orientation of the eigenvectors.\hfill$\blacksquare$

\subsection{Proof of Theorem \ref{T:new_uniqueness}}\label{T:new_uniqueness_proof}

From~\eqref{E:quadratic_system} it follows that
\begin{equation}
	\bbC_{\bby,1}\!=\!\bbH \bbC_{\bbx,1} \bbH^\top = \bbH \bbV_{\bbx} \diag (\bblambda_1) \bbV_{\bbx}^\top \bbH^\top = \bbQ \diag (\bblambda_1) \bbQ^\top\!\!,  \nonumber
\end{equation}
where we have implicitly defined $\bbQ = \bbH \bbV_{\bbx}$. Notice that the basis $\bbV_{\bbx}$ is completely determined (up to signs) since all eigenvalues in $\bblambda_1$ are distinct [cf. cond. i)]. Similarly, we obtain that $\bbC_{\bby,2} = \bbQ \diag (\bblambda_2) \bbQ^\top$. Alternatively, we get that $\bbC_{\bby,3} = \bbP \diag (\bblambda_3) \bbP^\top$ and $\bbC_{\bby,4} = \bbP \diag (\bblambda_4) \bbP^\top$ where $\bbP = \bbH \bbW_{\bbx}$. We define the matrices $\bbR_{\bbx,1} = [\bblambda_{1}, \bblambda_2]^\top$ and $\bbR_{\bbx,2} = [\bblambda_{3}, \bblambda_4]^\top \in \reals^{2 \times N}$.
Furthermore, we consider the $N \times N \times 2$ tensors $\underline{\bbC}_{\bby,1}$ and $\underline{\bbC}_{\bby,2}$ where the slices among the third mode in $\underline{\bbC}_{\bby,1}$ are given by $\bbC_{\bby,{1}}$ and $\bbC_{\bby,{2}}$ and those of $\underline{\bbC}_{\bby,2}$ are given by $\bbC_{\bby,{3}}$ and $\bbC_{\bby,{4}}$.
Given the introduced tensors, the partial symmetric PARAFAC decomposition of $\underline{\bbC}_{\bby,1}$ factors into matrices $\bbQ$, $\bbQ$, and $\bbR_{\bbx,1}$ and that of $\underline{\bbC}_{\bby,2}$ factors into $\bbP$, $\bbP$, and $\bbR_{\bbx,2}$; see~\cite{parafac, shen2017tensors}.

Recall that the Kruskal rank of a matrix $\bbA \in \reals^{N \times M}$ (denoted by $\mathrm{kr}(\bbA)$) is defined as the maximum number $k$ such that any combination of $k$ columns of $\bbA$ constitute a full rank submatrix. 
In this way, from condition ii) it follows that $\mathrm{kr}(\bbR_{\bbx,1}) = \mathrm{kr}(\bbR_{\bbx,2}) = 2$ and since $\bbH$ is full rank [cf. cond. iv)] it follows that $\mathrm{kr}(\bbQ) = \mathrm{kr}(\bbP) = N$. 
Leveraging established results on the uniqueness of PARAFAC tensor decompositions (see~\cite[Theorem 1]{shen2017tensors}), it follows that a PARAFAC decomposition of $\underline{\bbC}_{\bby,1}$ recovers $\bbQ$ and $\bbR_{\bbx,1}$ up to scaling and rotation ambiguities and a similar result holds for the decomposition of $\underline{\bbC}_{\bby,2}$. 
However, given that we know $\bbR_{\bbx,1}$ and $\bbR_{\bbx,2}$ a priori, part of those ambiguities can be resolved; see, e.g.,~\cite[Lemma 1]{shen2017tensors}. 
To be more precise, it follows that from the decomposition of $\underline{\bbC}_{\bby,1}$ we can recover $\bbQ'$, where $\bbQ' = \bbQ \diag(\bar{\bbb}_1)$ for some unknown $\bar{\bbb}_1 \in \{-1,1\}^{N}$. In a similar fashion, from the decomposition of  $\underline{\bbC}_{\bby,2}$ we can recover $\bbP'$, where $\bbP' = \bbP \diag(\bar{\bbb}_2)$ for some unknown $\bar{\bbb}_2 \in \{-1,1\}^{N}$
However, the ensuing lemma establishes how to uniquely recover $\bar{\bbb}_1$ and $\bar{\bbb}_2$.
\begin{mylemma}\label{lem:b}
	Vectors $\bar{\bbb}_1,\bar{\bbb}_2 \in \{-1,1\}^N$ can be found as the only vectors (up to sign) such that $\bbQ' \diag(\bar{\bbb}_1) \bbV_{\bbx}^\top = \bbP' \diag(\bar{\bbb}_2) \bbW_{\bbx}^\top$.	
\end{mylemma}
\begin{myproof}
	Since $\bbQ' \diag(\bar{\bbb}_1) \bbV_{\bbx}^\top = \bbH = \bbP' \diag(\bar{\bbb}_2) \bbW_{\bbx}^\top$, for the true $\bar{\bbb}_1,\bar{\bbb}_2$ the statement of the lemma readily holds. 
	Hence, we are left to show that no other signed binary vectors  ${\bbb}'_1$ and ${\bbb}'_2$ lead to $\bbQ' \diag({\bbb}'_1) \bbV_{\bbx}^\top = \bbP' \diag({\bbb}'_2) \bbW_{\bbx}^\top$.
	To do this, notice that from the definitions of $\bbQ'$ and $\bbP'$ we have that $\bbQ' = \bbH \bbV_{\bbx} \diag(\bar{\bbb}_1)$ and $\bbP' = \bbH \bbW_{\bbx} \diag(\bar{\bbb}_2)$. Thus, we may rewrite the sought equality as 
	\begin{align}
		\bbH \bbV_{\bbx} \diag(\bar{\bbb}_1) \diag({\bbb}'_1) \bbV_{\bbx}^\top = \bbH \bbW_{\bbx} \diag(\bar{\bbb}_2) \diag({\bbb}'_2) \bbW_{\bbx}^\top. \nonumber
	\end{align}
	Leveraging condition iv), this implies that 
	\begin{align}
		\bbV_{\bbx} \diag(\bar{\bbb}_1) \diag({\bbb}'_1) \bbV_{\bbx}^\top = \bbW_{\bbx} \diag(\bar{\bbb}_2) \diag({\bbb}'_2) \bbW_{\bbx}^\top. \nonumber
	\end{align}
	Finally, from condition iii), the above equality only holds when both sides are equal to $\bbI_N$ or $-\bbI_N$. 
	Hence, we must have that $\bbb'_1 = \bar{\bbb}_1$ and $\bbb'_2 = \bar{\bbb}_2$ or $\bbb'_1 = - \bar{\bbb}_1$ and $\bbb'_2 = - \bar{\bbb}_2$, from where identifiability of $\bar{\bbb}_1,\bar{\bbb}_2$ (up to a sign) is guaranteed.		
\end{myproof}

Finally, since Lemma~\ref{lem:b} guarantees that we can recover $\bar{\bbb}_1$ up to a sign, we get uniqueness (up to sign ambiguity) of $\bbH = \bbQ \bbV_{\bbx}^\top = \bbQ' \diag(\bar{\bbb}_1) \bbV_{\bbx}^\top$, as wanted.\hfill$\blacksquare$

\bibliographystyle{IEEEtran}
%
\bibliography{citations}


\newpage


\section*{Supplementary Material}\label{S:supp_material}



\subsection{Proof of Theorem \ref{T:alg_convergence}}\label{T:supp_proof}

Throughout, we collect the optimization variables in $\bbZ:=\{\bbS,\bbH,\{\bbU_m\}_{m=1}^{M}\}$ and for notational convenience we abbreviate the symmetric matrices
\begin{equation}\label{E:XY_def}
	\bbY_m:=\hbC_{\bby,m}^{1/2},\qquad \bbX_m:=\bbC_{\bbx,m}^{-1/2}.
\end{equation}
Consolidating the constraints of \eqref{E:joint_penalized} into the objective, we study the minimization of the proper, lower-semicontinuous
\begin{equation}\label{E:F_consolidated}
	F(\bbZ)=h(\bbZ)+g_{\bbS}(\bbS)+g_{\bbU}(\{\bbU_m\}),
\end{equation}
where $h$ is the $C^\infty$ polynomial objective in \eqref{E:filter_subproblem}, $g_{\bbS}(\bbS)=\|\bbS\|_1+\iota_{\ccalS}(\bbS)$, and $g_{\bbU}(\{\bbU_m\})=\sum_{m=1}^{M}\iota_{\ccalU_N}(\bbU_m)$; the $\bbH$ block carries no nonsmooth term. All variables are real and $\ccalU_N$ is the Stiefel manifold, a real-algebraic variety. With $\bbPhi:=\bbH\bbS-\bbS\bbH$, the Euclidean gradients of $h$ are [cf.~\eqref{eq:grad_g_modified},~\eqref{e:nabla_fm},~\eqref{eq:gradient_wrt_filter_jointproblem}]
\begin{align}
	\nabla_{\bbS} h &= \mu(\bbH^\top\bbPhi-\bbPhi\bbH^\top), \label{E:gradS}\\
	\nabla_{\bbH} h &= \mu(\bbPhi\bbS^\top-\bbS^\top\bbPhi) \nonumber\\
	&\quad +\lambda\textstyle\sum_{m=1}^M(\bbH-\bbY_m\bbU_m\bbX_m), \label{E:gradH}\\
	\bbG_m:=\nabla_{\bbU_m} h &= \lambda\,\bbY_m(\bbY_m\bbU_m\bbX_m-\bbH)\bbX_m. \label{E:gradU}
\end{align}
The Riemannian gradient on $\ccalU_N$ is the projection of $\bbG_m$ onto the tangent space $\mathrm{T}_{\bbU_m}\ccalU_N$ [cf.~\eqref{e:grad_fm}],
\begin{equation}\label{E:riem_grad_supp}
	\grad_{\bbU_m} h=\bbG_m-\bbU_m\,\mathrm{sym}(\bbU_m^\top\bbG_m),
\end{equation}
where $\mathrm{sym}(\bbM):=\tfrac12(\bbM+\bbM^\top)$. The $\bbU_m$ blocks use the polar retraction $R_{\bbU}(\bbxi)=\mathrm{polar}(\bbU+\bbxi)=\bbP\bbQ^\top$ for tangent $\bbxi$, where $\bbP\bbSigma\bbQ^\top=\textrm{svd}[\bbU+\bbxi]$  [cf.~\eqref{eq:r_grad_step}]~\cite[p. 161]{boumal2023intromanifolds}.

\begin{mylemma}[Compact sublevel set]\label{L:compact}
	Under \textnormal{A1-A2}, $F$ is coercive and the sublevel set $\ccalC_0:=\{\bbZ:F(\bbZ)\leq F(\bbZ^0)\}$ is compact.
\end{mylemma}
\begin{myproof}
	$\ccalU_N$ is compact, so the $\bbU_m$ coordinates of any sublevel set are bounded. Since $F(\bbZ)\geq\|\bbS\|_1$, $F$ is coercive in $\bbS$. Because $\lambda>0$, $\{\bbU_m\}$ are bounded, and $\bbY_m,\bbX_m$ are fixed, $\frac{\lambda}{2}\sum_m\|\bbH-\bbY_m\bbU_m\bbX_m\|_F^2\to\infty$ as $\|\bbH\|_F\to\infty$, so $F$ is coercive in $\bbH$. Thus $\ccalC_0$ is closed and bounded.
\end{myproof}

\noindent\textbf{Explicit constants and step sizes making A3 precise.} Fix a compact set $\ccalC\supseteq\ccalC_0$. Since $h\in C^\infty$, on $\ccalC$ the partial gradients $\nabla_{\bbS} h,\nabla_{\bbH} h$ are Lipschitz with moduli $L_{\bbS},L_{\bbH}$, the full gradient $\nabla h$ is Lipschitz with modulus $L$, and each $\grad_{\bbU_m} h$ is Lipschitz with modulus $L_{\bbU}'$ and bounded by $\Gamma:=\sup_{\bbZ\in\ccalC_0,m}\|\grad_{\bbU_m} h\|_F<\infty$. Concretely, since $\nabla_{\bbS} h$ stems only from the commutator term $\bbPhi$, one may take $L_{\bbS}=4\mu\,\sigma_{\max}^2(\bbH)$ on $\ccalC_0$. A matching bound $L_{\bbH}\leq 4\mu\,\sigma_{\max}^2(\bbS)+\lambda M$ holds for the $\bbH$-block. The pullback $\bbxi\mapsto h(R_{\bbU_m}(\bbxi))$ has $\widehat L_{\bbU}$-Lipschitz gradient (retraction descent lemma~\cite{BAC2019}). Since the polar retraction is smooth with $DR_{\bbU}(\bbzero)=\mathrm{Id}$~\cite{boumal2023intromanifolds}, compactness of $\ccalU_N$ yields constants $\gamma_{\bbU},\kappa_{\bbU}>0$ and a radius $r_0>0$ with
\begin{equation}\label{E:two_sided_retr}
	\tfrac{1}{\kappa_{\bbU}}\|\bbxi\|_F\leq\|R_{\bbU}(\bbxi)-\bbU\|_F\leq\gamma_{\bbU}\|\bbxi\|_F
\end{equation}
for all $\bbU\in\ccalU_N$ and tangent $\bbxi$ with $\|\bbxi\|_F\leq r_0$. We then fix
\begin{equation}\label{E:stepsizes}
	\alpha_{\bbS}\in(0,1/L_{\bbS}),\ \ \alpha_{\bbH}\in(0,1/L_{\bbH}),\ \ \alpha_{\bbU}\in(0,2/\widehat L_{\bbU}),
\end{equation}
the last with $\alpha_{\bbU}\Gamma\leq r_0$, so each tangent step $\bbxi_m=-\alpha_{\bbU}\grad_{\bbU_m} h$ obeys $\|\bbxi_m\|_F\leq r_0$ and \eqref{E:two_sided_retr} holds.\vspace{2pt} 

\noindent\textbf{Stationarity.} As $h\in C^1$ and $g_{\bbS},g_{\bbU}$ are block-separable, the limiting subdifferential separates blockwise; for the $\bbU_m$ block, the distance from $0$ to the (normal-cone) subdifferential equals $\|\grad_{\bbU_m} h\|_F$, the norm of the tangential component of $\bbG_m$~\cite{boumal2023intromanifolds}. Hence $\mathbf{0}\in\partial F(\bbZ^\star)$ is equivalent to prox-stationarity of $\bbS^\star$, $\nabla_{\bbH} h(\bbZ^\star)=\bbzero$, and $\grad_{\bbU_m} h(\bbZ^\star)=\bbzero$ for all $m$, i.e., the critical-point notion stated before Theorem~\ref{T:alg_convergence}.

We have all the ingredients to establish our main convergence result.\vspace{2pt}

\begin{myproof}
Let $T:=P_{\bbS}+M P_{\bbU}+P_{\bbH}$ be the number of inner (single-block) steps per outer cycle ($P_\bullet=t_{\max}$ in Algorithm~\ref{alg:topoid_directed_inexact}, but the number of inner updates per block need not coincide). Index the inner increments of cycle $k$ by $\bbDelta_{k,1},\dots,\bbDelta_{k,T}$ and set $\delta_k:=\sum_{i=1}^{T}\|\bbDelta_{k,i}\|_F$, so $\|\bbZ^{k+1}-\bbZ^k\|_F\leq\delta_k$. The argument proceeds in six steps.\vspace{2pt}

\noindent\emph{Step 1 (Boundedness).} By Lemma~\ref{L:compact}, $\ccalC_0$ is compact; the per-step descent of Step~2 below keeps all iterates in $\ccalC_0$, where the constants above are valid.\vspace{2pt}

\noindent\emph{Step 2 (Sufficient decrease).} For an inner $\bbS$-step $\bbS^{+}=\mathrm{prox}_{\alpha_{\bbS} g_{\bbS}}(\bbS-\alpha_{\bbS}\nabla_{\bbS} h)$, the proximal operator definition and the descent lemma for $h$ give a decrease $\geq c_{\bbS}\|\bbS^{+}-\bbS\|_F^2$ with $c_{\bbS}=\frac{1}{2\alpha_{\bbS}}-\frac{L_{\bbS}}{2}>0$ (convexity of $g_{\bbS}$ only makes $\bbS^{+}$ unique). An inner $\bbH$-step is identical, with $c_{\bbH}=\frac{1}{2\alpha_{\bbH}}-\frac{L_{\bbH}}{2}>0$. For a $\bbU_m$-step \eqref{eq:r_grad_step} along $\bbxi=-\alpha_{\bbU}\grad_{\bbU_m} h$, the retraction descent lemma~\cite{BAC2019} yields
\begin{equation*}
\begin{aligned}
h(\dots,\bbU_m^{+},\dots)\leq{}& h(\dots,\bbU_m,\dots)\\
&-\alpha_{\bbU}\left(1-\tfrac{\widehat L_{\bbU}\alpha_{\bbU}}{2}\right)\|\grad_{\bbU_m} h\|_F^2,
\end{aligned}
\end{equation*}
and, as the step stays on $\ccalU_N$, $g_{\bbU}$ is unchanged so the bound holds for $F$. The upper bound in \eqref{E:two_sided_retr} gives $\|\grad_{\bbU_m} h\|_F\geq\frac{1}{\gamma_{\bbU}\alpha_{\bbU}}\|\bbU_m^{+}-\bbU_m\|_F$, whence the decrease is $\geq c_{\bbU}\|\bbU_m^{+}-\bbU_m\|_F^2$ with $c_{\bbU}=\frac{1-\widehat L_{\bbU}\alpha_{\bbU}/2}{\gamma_{\bbU}^2\alpha_{\bbU}}>0$. With $a:=\min\{c_{\bbS},c_{\bbH},c_{\bbU}\}>0$, summing over the $T$ inner steps and using Cauchy--Schwarz,
\begin{equation}\label{E:H1}
	F(\bbZ^k)-F(\bbZ^{k+1})\geq a\sum_{i=1}^{T}\|\bbDelta_{k,i}\|_F^2\geq\tfrac{a}{T}\,\delta_k^2,
\end{equation}
establishing part~(i).\vspace{2pt}

\noindent\emph{Step 3 (Summability).} Telescoping \eqref{E:H1}, $\tfrac{a}{T}\sum_{k\geq0}\delta_k^2\leq F(\bbZ^0)-F^\star<\infty$, so $\delta_k\to0$ and $\|\bbZ^{k+1}-\bbZ^k\|_F\to0$.\vspace{2pt}

\noindent\emph{Step 4 (Relative error).} We build $\bbW^{k+1}\in\partial F(\bbZ^{k+1})$ with $\|\bbW^{k+1}\|_F\leq b\,\delta_k$, blockwise. For the $\bbS$ block, let the last inner $\bbS$-step start at base point $\tilde\bbZ$; its prox optimality reads $\tfrac{1}{\alpha_{\bbS}}(\bbS^{-}-\bbS^{k+1})-\nabla_{\bbS} h(\tbZ)\in\partial g_{\bbS}(\bbS^{k+1})$, so adding $\nabla_{\bbS} h(\bbZ^{k+1})$ yields an element of $\partial_{\bbS}F(\bbz^{k+1})$ of norm $\leq(\tfrac{1}{\alpha_{\bbS}}+L)\delta_k$, using $\|\bbZ^{k+1}-\tbZ\|_F\leq\delta_k$. Analogously for the $\bbH$ block, with bound $(\tfrac{1}{\alpha_{\bbH}}+L)\delta_k$. Finally for the $\bbU_m$ block, the residual is $\|\grad_{\bbU_m} h(\bbZ^{k+1})\|_F$; with $\bbxi_m$ the last tangent step, the lower bound in \eqref{E:two_sided_retr} gives $\|\bbxi_m\|_F\leq\kappa_{\bbU}\|\bbU_m^{k+1}-\bbU_m^{-}\|_F$, hence $\|\grad_{\bbU_m} h(\tbZ)\|_F=\tfrac{1}{\alpha_{\bbU}}\|\bbxi_m\|_F\leq\tfrac{\kappa_{\bbU}}{\alpha_{\bbU}}\|\bbU_m^{k+1}-\bbU_m^{-}\|_F$, and $L_{\bbU}'$-Lipschitzness of $\grad_{\bbU_m} h$ yields $\|\grad_{\bbU_m} h(\bbZ^{k+1})\|_F\leq(L_{\bbU}'+\tfrac{\kappa_{\bbU}}{\alpha_{\bbU}})\delta_k$. Stacking, we find
\begin{equation}\label{E:H2}
	\mathrm{dist}\big(\mathbf{0},\partial F(\bbZ^{k+1})\big)\leq b\,\delta_k,
\end{equation}
with $b:=(M{+}2)\max\{\tfrac{1}{\alpha_{\bbS}}{+}L,\ \tfrac{1}{\alpha_{\bbH}}{+}L,\ L_{\bbU}'{+}\tfrac{\kappa_{\bbU}}{\alpha_{\bbU}}\}$.\vspace{2pt}

\noindent\emph{Step 5 (Subsequential stationarity).} Let $\bbZ^{k_j}\to\bar\bbZ$ (compactness); by Step~3, $\bbZ^{k_j+1}\to\bar\bbZ$. On the closed domain $\ccalS\times\reals^{N\times N}\times\ccalU_N^M$ the indicators vanish and $F$ equals the continuous $h+\|\cdot\|_1$, so $F(\bbZ^{k_j})\to F(\bar\bbZ)$; as $\{F(\bbZ^k)\}$ decreases to $F^\star$, $F(\bar\bbZ)=F^\star$. By \eqref{E:H2}, $\mathrm{dist}(\mathbf{0},\partial F(\bbZ^{k_j+1}))\to0$, and closedness of the graph of $\partial F$ gives $\mathbf{0}\in\partial F(\bar\bbZ)$, proving part~(ii).\vspace{2pt}

\noindent\emph{Step 6 (KL and finite length).} By A1-A2, $h$ is polynomial, $\|\cdot\|_1$ is semialgebraic, $\ccalU_N$ is a real-algebraic variety, and $\ccalS$ is semialgebraic [cf.~A1]; hence $F$ is semialgebraic and satisfies the KL property everywhere~\cite{ABS2013}. Inequalities \eqref{E:H1}--\eqref{E:H2} are the descent and relative-error hypotheses of the framework of~\cite{ABS2013,BST2014} (with increment surrogate $\delta_k$). Concavity of the KL desingularizing function $\varphi$ gives
\begin{equation*}
\begin{aligned}
&\psi\!\big(F(\bbZ^k){-}F^\star\big)-\psi\!\big(F(\bbZ^{k+1}){-}F^\star\big)\\
&\qquad\geq\ \tfrac{(a/T)\,\delta_k^2}{b\,\delta_{k-1}},
\end{aligned}
\end{equation*}
so $\delta_k^2\leq\tfrac{bT}{a}\delta_{k-1}(\psi_k-\psi_{k+1})$. Using the arithmetic mean-geometric mean inequality and telescoping gives $\sum_k\delta_k<\infty$. Thus, $\{\bbZ^k\}$ has finite length, is Cauchy, and converges to a single $\bbZ^\star$ with $\mathbf{0}\in\partial F(\bbZ^\star)$, establishing part~(iii).
\end{myproof}

\noindent\textbf{Convergence rate.} Having established the finite length of $\{\bbZ^k\}$, the exponent-dependent estimates of~\cite{ABS2013,BST2014}, applied to the tail length $\Delta_k:=\sum_{\ell\geq k}\|\bbZ^{\ell+1}-\bbZ^{\ell}\|_F\geq\|\bbZ^k-\bbZ^\star\|_F$, give: if $F$ has KL exponent $\theta\in[0,1)$ at $\bbZ^\star$, the iteration terminates finitely for $\theta=0$, converges linearly ($\Delta_k\leq C\rho^k$, $\rho\in(0,1)$) for $\theta\in(0,\tfrac12]$, and sublinearly ($\Delta_k=\ccalO(k^{-(1-\theta)/(2\theta-1)})$) for $\theta\in(\tfrac12,1)$.

\begin{remark}[Discussion and extensions]\normalfont The inner-iteration counts and the cyclic block order enter only through the constants $a/T$ and $b$, never the conclusions, so our analysis covers any finite $t_{\max}\geq1$ and any deterministic block schedule. A complementary, KL-free route to best-iterate rates for BCD on smooth manifolds is developed in~\cite{peng2023bcdmanifold}; extending it here is hindered by the nonsmooth term $\|\bbS\|_1+\iota_{\ccalS}$ on the $\bbS$ block, which we leave for future work.
\end{remark}

\end{document}

%% file: my_sections.tex
\usepackage{needspace}

